\documentclass[10pt,journal,compsoc]{IEEEtran}
\usepackage{graphicx}
\usepackage{amsmath}
\usepackage{commath}
\usepackage{algorithm2e}
\usepackage{ragged2e}
\usepackage{float}
\usepackage{dblfloatfix}
\usepackage{amssymb}
\usepackage{comment}
\usepackage{soul}
\usepackage{xcolor}
\usepackage{multirow}
\usepackage{varioref}
\usepackage{xr-hyper}
\usepackage{hyperref}
\hypersetup{%
	pdfborder = {0 0 0}
}
\usepackage{threeparttable}
\graphicspath{{Figures/}}
\ifCLASSINFOpdf
\else
\fi
\begin{document}
	
	\title{Learning Inter- and Intra-manifolds\\ for Matrix Factorization-based \\Multi-Aspect Data Clustering}
	
	\author{Khanh~Luong,
		Richi~Nayak
		\IEEEcompsocitemizethanks{\IEEEcompsocthanksitem K. Luong and R. Nayak are with the School of
			Computer Science and QUT Centre for Data Science, Queensland University of
			Technology, Australia, QLD, 4000.
			\protect\\
			E-mail: khanh.luong@qut.edu.au, r.nayak@qut.edu.au
		}
	} 

\markboth{Journal of \LaTeX\ Class Files,~Vol.~14, No.~8, August~2015}%
{Shell \MakeLowercase{\textit{et al.}}: Bare Demo of IEEEtran.cls for Computer Society Journals}

\IEEEtitleabstractindextext{%
	\begin{abstract}
		Clustering on the data with multiple aspects, such as multi-view or multi-type relational data, has become popular in recent years due to their wide applicability. The approach using manifold learning with the Non-negative Matrix Factorization (NMF) framework, that learns the accurate low-rank representation of the multi-dimensional data, has shown effectiveness. We propose to include the inter-manifold in the NMF framework, utilizing the distance information of data points of different data types (or views) to learn the diverse manifold for data clustering. Empirical analysis reveals that the proposed method can find partial representations of various interrelated types and select useful features during clustering. Results on several datasets demonstrate that the proposed method outperforms the state-of-the-art multi-aspect data clustering methods in both accuracy and efficiency.
	\end{abstract}
	
	}

\maketitle

\IEEEdisplaynontitleabstractindextext

\IEEEpeerreviewmaketitle

\section{Introduction}
\begin{figure*}
\includegraphics[width=0.7\linewidth]{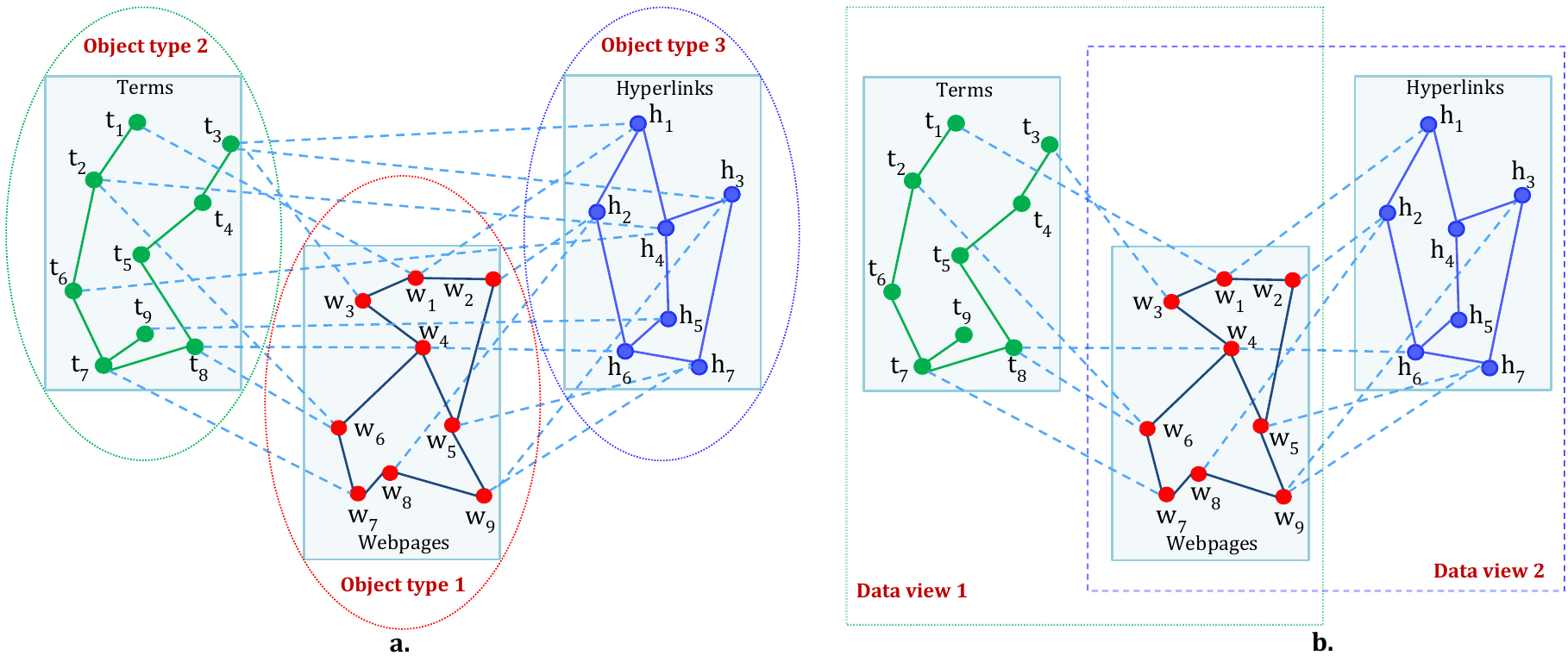}
\centering
\caption{Example of Multi-Aspect Data. Fig. \ref{fig:Multi_Aspect_Example}.a. A Multi-type Relational Data with three object types: Webpages, Terms, Hyperlinks. Fig. \ref{fig:Multi_Aspect_Example}.b. A Multi-view Data with two views: Terms and Hyperlinks. The intra-type/view relationships are represented as solid lines and inter-type/view relationships are represented as dotted lines.}
\label{fig:Multi_Aspect_Example}
\end{figure*}

\IEEEPARstart{M}{ulti-Aspect} data, the data represented with multiple aspects, are becoming common and useful in practice \cite{Long}. This can be (1) \textit{multi-view} data where samples are represented by multiple views; or (2) \textit{multi-type relational data (MTRD)} where samples are represented by different data types and their inherent relationships. Aspects in MTRD are different object types (together with their associated relationships) and in multi-view data are multiple views. An example of multi-aspect data is given in Figure \ref{fig:Multi_Aspect_Example}. Figure \ref{fig:Multi_Aspect_Example}.a. shows a MTRD dataset with 3 object types: Webpages, Terms and Hyperlinks with various relationships between objects of these object types. Figure \ref{fig:Multi_Aspect_Example}.b. shows an example of two-view data where the first view is the representation of Webpages by Terms and the second view is the representation of Webpages by Hyperlinks. The Webpages can be considered as samples while Terms and Hyperlinks are considered as features.  

The objective of MTRD data clustering is to make use of as many relationships as possible to cluster all object types simultaneously to enhance the clustering performance. In multi-view data clustering, the relationship between different feature objects (e.g., between Terms and Hyperlinks) is ignored and the objective is to cluster samples using all view data and look for the consensus cluster structure\footnote{In this paper, we will consider the context of MTRD and relate to multi-view data when necessary.} \cite{Luong2019}. 

MTRD is represented by multiple types of objects and two main types of relationships, namely \textit{inter-type} and \textit{intra-type} \cite{Wang:2011}. Inter-type relationships contain connections between objects from different types (e.g., co-occurrences between Webpages and Terms) and intra-type relationships contain the connections between objects of the same type (e.g., similarities between Webpages). Each intra-type or inter-type relationship is normally encoded by a matrix. For instance, in Figure \ref{fig:Multi_Aspect_Example}.a, three intra-type relationship matrices will store intra-similarities between Webpages, between Terms and between Hyperlinks, and three inter-relationships matrices will store relationships between Webpages and Terms, between Webpages and Hyperlinks and between Terms and Hyperlinks. It is conjectured that the inclusion of all object types and their relationships in clustering will provide a detailed view of data and will yield an accurate and informative clustering solution.

The MTRD data is normally very sparse and high dimensional. A concatenated matrix combining all multiple relationships fails to include the inter-relatedness between the data and results in inferior performance \cite{Luong2019}.
This fact leads most existing state-of-the-art MTRD methods to learn a low-rank representation\footnote{The terms low-rank and low-dimensional have been used interchangeably.} of the data and apply clustering on the newly found representation (i.e., reduced features set). Subspace learning \cite{Subspace_Review:2004}\cite{Tutorial_subspace} and Non-negative Matrix Factorization (NMF) \cite{Lee:1999}\cite{NMF2001} are two well-known dimensionality reduction approaches. While subspace learning-based methods focus on feature selection, they do not support the feature transformation from higher to lower order \cite{Subspace_Review:2004}. NMF decomposes the original matrix into low-rank matrices that embed the original latent structure. It has been shown to learn the good part-based representation for sparse datasets thanks to the non-negativity constraint \cite{Lee:1999}. NMF-based clustering methods have been known to outperform subspace clustering methods on high dimension, high relatedness datasets such as text-based data \cite{ASurveyofTextClustering:2012}.

The real-world data is known to lie on multiple manifolds embedded in the high and multi dimensional data space \cite{MMNMF:2017}\cite{Belkin:2003}\cite{Khanh}. Manifold learning has been applied to preserve the geometric structure of original data when projecting from high to lower dimensional space \cite{Belkin:2006}. Manifold learning has been incorporated in NMF to learn the part-based representation that respects the intrinsic manifold of data and assists in obtaining more accurate clusters \cite{Luong2019}\cite{Cai:2011}\cite{Gu:2009}\cite{TKDE}. 

Many efforts have been made to learn the accurate manifold of MTRD \cite{Wang:2011}\cite{Khanh}\cite{Li:2013}\cite{Jun}. These methods focus on learning the manifold on each object type explicitly, i.e., considering how data points of the same object type reside on the manifold. They ignore the intrinsic relatedness between the data of multiple types and learn an incomplete manifold. The sampling of data points from different types should follow an intrinsic shape (or manifold) due to the high relatedness between them. We call this process \textit{inter-manifold learning} and differentiates it from \textit{intra-manifold learning} which focuses on the sampling of data points within the same data type on a manifold. To the best of our knowledge, no method exists that exploits the residing of data points from different types on a manifold (e.g., sampling on a manifold of Webpages and Terms).  

We propose to construct a $p$ nearest neighbour ($p$NN) graph for each inter-relationship to capture the closely related objects from two different types and aim to maintain this closeness during the low-order representation generation and clustering. The aim is to accurately learn multiple manifolds generated on both intra-relationships and inter-relationships of data to capture all different relationships, and steadily preserve the learned diverse manifold when mapping to the new low-dimensional data space with NMF. We propose a novel NMF framework to regularize the complete and diverse geometric information in multi-aspect data. 
We call the proposed method as clustering with \textit{Di}verse \textit{M}anifold for \textit{M}ulti-\textit{A}spect Data (DiMMA). 

The low-rank representation learned from DiMMA can be used for various purposes such as information retrieval, classification and clustering, especially showing effectiveness in the context when data is presented with multiple aspects. We conducted rigorous experiments to test the clustering performance of DiMMA on multi-aspect data and benchmarked it with relevant NMF-based methods. Empirical analysis reveals that DiMMA can cluster multi-aspect data accurately and efficiently. Distinct from existing methods \cite{Wang:2011}\cite{Li:2013}\cite{Jun}\cite{Zong:2017}, DiMMA can simultaneously find a low-rank representation of various interrelated object types such that the distance relationships between highly related objects from the same and different types are incorporated in the low-dimensional projected space. This is the first contribution of this paper that assists DiMMA in producing effective outcomes.

Another contribution of DiMMA lies in its ability to identify and retain the most important features during the clustering process. This does not only help in significantly improving the clustering performance but also aids in dealing with high dimensionality, a problem inherent in multi-aspect data. We conducted an experiment to investigate the capability of DiMMA in selecting good features compared to its counterparts. The cohesiveness of a cluster solution, the sum of the average cohesiveness of each sample to its cluster, is calculated.

Lastly, we compare DiMMA with the state-of-the-art Deep Neural Network (DNN) based approach to learning a low-rank representation. The experimental result shows that the NMF-based DiMMA is able to learn the cluster structures better than the DNN-based approach in datasets with small and medium sizes.  

In summary, we formulate and solve the problem of clustering multi-aspect data by simultaneously learning from the input data matrices directly, rather than relying on reformulating a big symmetric matrix \cite{Wang:2011} as the common approach of existing MTRD clustering methods \cite{Li:2013}\cite{Jun}.  
For the newly formulated objective function, we offer a set of update rules with a guarantee on the correctness and convergence of DiMMA. Extensive experiments ascertain its effectiveness in obtaining an accurate clustering solution.

The rest of the paper is organized as follows. Section \ref{sec:relatedworks} details the related works on low-rank representation learning methods.
Sections \ref{sec:method} and \ref{sec:experiment} present DiMMA and its empirical analysis. The conclusion and future work are in Section \ref{sec:conclusion}.

\section{Related work}
\label{sec:relatedworks}
In this paper, we present the data for each view/type as a matrix data and use a NMF based method to explore the relationships present in the data for identifying clusters. The multi-aspect data can also be presented as a heterogeneous information network where multiple nodes represent objects of different object types. The problem of a heterogeneous information network clustering is approached either by simultaneously clustering objects of all types or clustering the target-objects based on attribute objects \cite{CluteringonHeterogeneousNetwork:2014}. For example, RankClus \cite{RankClus:2009} and PathSeClus \cite{PathSeClus:2013} cluster target objects (sample objects) by considering the relationship to other object types as attributes or features which may be useful only for the network with an object type in the centre.

Although these methods can handle the heterogeneity of networks, the sparseness issue remains unsolved \cite{NMF_for_CommunityDetection:2015}\cite{tang2012community}. Moreover, these networks focus on representing different types of relationships among the same type of objects (e.g., user-user). They ignore the relationships among different types of objects (e.g., user-hashtag, user-term) that can be significant for grouping objects (e.g., users) based on similar interest. A recent work \cite{MTRD_Community:2019} has compared the NMF based methods and networks methods to exploit different types of information in social network and have found the NMF based MTRD clustering method more useful dealing with sparse data.

\subsection{NMF Clustering and Manifold Learning }

NMF clustering \cite{Cai:2011} is an established method for high-dimensional and sparse data to obtain a new mapped low-dimension space and search clusters in the new space. NMF-based methods can perform both hard and soft clustering as well as the outputs of NMF can easily be extended to various data types due to the nature of matrix factorization.
Early works that extend NMF \cite{NMF2001} to two-type data \cite{Ding:2006} and to multi-type data \cite{Long} produce insufficient accuracy since they did not consider the local geometric structure which is effective for identifying clusters. However, the later works that use geometric information on two-type data \cite{Gu:2009}\cite{Li:2013}, multi-type data \cite{Wang:2011}\cite{Khanh}\cite{Jun}, and multi-view \cite{Zong:2017}\cite{MVobjectRecognition:2018} have achieved significant improvement. 

Manifold learning is long known for non-linear dimensionality reduction by preserving the local and/or global geometric structure of original data to the projected space. 
Manifold learning, also referred as graph embedding, considerably differs from network embedding \cite{AsurveyonNetworkEmbedding:2019}. Network embedding usually refers to the networks that are formed naturally as in social networks and biology networks whereas in manifold learning, a graph is constructed from the dataset feature representation. In other word, network embedding works on the first-order (i.e., original) network to seek for a useful network embedding for the purpose of network inference, whereas manifold learning works on the second-order graph to maintain the embedded structure of the high-dimensional data for learning a more accurate low-dimensional representation. 

In the last decade, researchers have combined NMF with manifold learning to obtain a representative mapped low-dimension space that encodes the geometric structure of the data explicitly  \cite{Wang:2011}\cite{Cai:2011}\cite{Li:2013}\cite{Zong:2017}. This family of methods uses a similarity graph to encode the local geometry information of the original data and to discretely approximate the manifold so the sampling of data on the manifold can be investigated \cite{Huang:2014robust}. 

A similarity graph, to model the local neighbourhood relationships between data points, is usually built by using two distance-based concepts, i.e., $\epsilon$-neighbourhood and $k$-nearest neighbour \cite{Luxburg:2007}. By incorporating the NN graph as an additional regularized term in NMF when projecting the data to low-dimensional space, the close distances of data points are maintained during the learning process \cite{Belkin:2006}. This corresponds to maintaining the high relatedness between points, in other words,  maintaining the local structure/shape of the original data. In the manifold theorem, this corresponds to optimally learning and preserving the intrinsic manifold of the original data in a classification or clustering problem \cite{Belkin:2003}.

It continues to be a non-trivial task to learn the accurate manifold of multi-aspect data. RHCHME \cite{Jun} tries to learn the accurate and complete intra-type relationships by considering the data lying on the manifold. In \cite{Zong:2017}, by generating the intrinsic manifold of a multi-view data embedded from a convex hull of all the views' manifolds, the consensus manifold is learned via a linear combination of multiple manifolds of data from all views. 
The consensus manifold is further exploited by the novel optimal manifold concept proposed in 2CMV \cite{KhanhICDE2020} which embeds the most consensed manifold in multi-view data. Recently, ARASP \cite{Khanh} proposed a novel fashion of learning the MTRD manifold, where the close and far distance information is embedded and preserved steadily for each data type. These works have proved the importance and necessity of manifold learning with clustering. 

However, these manifold-based methods mainly focus on using the geometric structure of each object type in MTRD (or each view in multi-view data) only. They fail to consider the sampling on manifolds of data points from different object types in MTRD or fail to consider the sampling on manifolds of data samples and data features in multi-view data. By ignoring the local geometric structure in multi-aspect data, these methods result in poor accuracy. This paper proposes to exploit all geometric structure information from the original data graph and aims to use the relationships within each object type or view as well as the relationships from different data types or views. 

\subsection{Non NMF-based Multi-Aspect Data Clustering}
There exist other non NMF-based approaches to solving the problem of clustering multi-aspect data such as spectral-based \cite{Long}\cite{IterativeViewsAgreement:2016}\cite{Co-regularized:2011} or subspace-based clustering \cite{Subspace_Review:2004}\cite{Tutorial_subspace}. Spectral-based methods, which is based on the well-studied spectral graph theory, aim to partition the similarity data graph to find its best cut responding to a clustering problem. The graph partitioning process is conducted using eigenvectors of Laplacian graph built from the data affinity graph \cite{Luxburg:2007}. Solving the eigenvector problem requires high computational cost, hence, this spectral-based methods group is known to perform poorly with the high dimensional data \cite{Dhillon:2005}. Subspace-based methods, a counterpart of NMF-based methods, seek a low-rank representation by selecting features from the original high dimensional data. However, this family of methods struggles to learn a good feature transformation due to not retaining the relations present in original data when projecting to a lower-order space for the clustering task \cite{ASurveyofTextClustering:2012}. 

In parallel, this decade has witnessed a surge in the development of deep learning-based methods for data learning, ranging from using the autoencoder, using feed-forward networks or using Generative Adversarial Network (GAN) \cite{DeepLearning_Survey:2015}\cite{ AsurveyofClusteringWithDeep:2018}. However, limited research has been done to exploit deep learning in clustering multi-aspect data. 

The auto-encoder technique, where a compact low-rank representation is learned between the coding and decoding process, has been exploited for traditional one-aspect data clustering \cite{ Autoencoder-basedC:2013} and multi-aspect data learning \cite{Ngiam:2011}. The multi-modal deep learning \cite{Ngiam:2011} is a prominent method wherein the shared representation between input modalities has been learned and used for different purposes such as reconstructing each or both modalities or to use on the test data for classification tasks. Another group of DNN-based methods uses canonical correlation analysis (CCA) to constrain data learned from different views while using DNN \cite{DeepLearning_Survey:2015}. DCCA \cite{DCCA:2013} is the most noticeable method that exploited CCA in DNN for learning two-view data. A distinct network is used to learn the low-rank compact data for each view and the two low-rank representations are forced to be maximally correlated. DCCAE \cite{DCCAE:2015} adds an auto-encoder-like constraint to DCCA and can maximally reconstruct the original data. 

There has been substantial research in DNN-based representation learning, however, these methods have not been fully exploited in multi-aspect data clustering specifically as well as they require the availability of training data. An emerging direction that bridges the unsupervised and supervised approaches has been reported in recent works of transfer learning \cite{UnsupervisedDeep:2016}\cite{JointUnsupervised:2016}\cite{DeepClustering:2018}. These works use an unsupervised learning method to learn the label for a supervised learning method. This novel approach is expected to work for a wide range of applications where the large datasets are available with the lack of labeled information, however, this approach is yet to be designed for multi-aspect data. 

\section{DiMMA: Learning Diverse Manifold for Multi-Aspect Data Clustering}
\label{sec:method}
\subsection{Definitions and Notations}
\subsubsection{Multi-type Relational Data}
Let $D = \{X_{1}, X_{2}, ..., X_{m}\}$ be the dataset with $m$ object types. Each object type $X_{h} = \{x_{i}\}_{(1\leq i \leq n_{h})}$ is a collection of $n_{h}$ objects.

\subsubsection{Inter-type relationship}
The inter-type relationship (or inter-relationship for simplicity) represents the relation between objects of different object types. The inter-type relationship matrix $R_{hl}$ representing relationships between objects of object type $X_h$ and object type $X_l$, is denoted as below,  

\begin{equation} 
R_{hl}=
\begin{bmatrix} r_{11} &r_{12} &\cdots &r_{1n_l}
\\ r_{21} &r_{22} &\cdots &r_{2n_l}
\\ \cdots &\cdots &\cdots &\cdots
\\ r_{n_h1} &r_{n_h2} &\cdots &r_{n_hn_l}
\end{bmatrix}
\end{equation}
where $r_{ij}$ is the relationship of object $i$th of object type $X_h$ and object $j$th of object type $X_l$. For example, suppose $X_h$ and $X_l$ are the data type Document and Term, respectively, then $r_{ij}$ denotes the tf-idf weight of a term in a document. 
\subsubsection{Intra-type relationship}
The intra-type relationship (or intra-relationship for simplicity) represents the relation between objects of the same object type. The intra-relationship matrix $W_{h}$ representing relationships among objects of the same object type $X_h$, is denoted as below,  

\begin{equation} 
W_{h}=
\begin{bmatrix} w_{11} &w_{12} &\cdots &w_{1n_h}
\\ w_{21} &w_{22} &\cdots &w_{2n_h}
\\ \cdots &\cdots &\cdots &\cdots
\\ w_{n_h1} &w_{n_h2} &\cdots &w_{n_hn_h}
\end{bmatrix}
\label{eq:intrarelationship}
\end{equation}
$W_{h} = \{w_{ij}\}^{n_{h}\times n_{h}}$ is a weighted adjacency matrix resulted from building a $k$ nearest neighbour graph ($k$NN graph) of object type $X_h$ \cite{Wang:2011}\cite{Cai:2011}\cite{Jun}. For example, suppose $X_h$ is the data type Document then $w_{ij}$ denotes the affinity between documents $i$th and $j$th in the dataset. 

\subsubsection{MTRD Clustering Problem}
Let $R_{hl}\in \{R_+\}_{1\leq h \neq l \leq m}$ be a set of inter-relationship matrices where $R_{hl}=R^{T}_{lh}$ and 
$\{W_{h}\}_{1\leq h \leq m}$ be a set of intra-relationship matrices of $m$ object types generated for the data $D = \{X_{1}, X_{2}, ..., X_{m}\}$. The task of clustering is to simultaneously group $m$ object types in $D$ into $c$ clusters by using the inter-type relationships between objects of different types, i.e., $\{R_{hl}\}$ as well as considering the intra-type relationships between objects in the same type, i.e., $\{W_{h}\}$. 
We assume that the objects in the dataset, which represent different types of features, tend to have the same number of underlying groups due to their high inter-relatedness. This assumption allows us to set the same dimensionality for the newly mapped low-dimensional space of all object types. 

The NMF-based objective function for MTRD is as \cite{Long}\cite{Wang:2011},
\begin{equation} J_{1} = \min \sum_{1\leq h<l \leq m}\|R_{hl} - G_{h}S_{hl}G^{T}_{l}\|^{2}_{F} \mbox{, } G_{h}\geq 0, G_{l}\geq 0 
\label{eq:J1}
\end{equation}
where $\|.\|_{F}$  denotes the Frobenius norm, $G_{h}$ and $G_{l}$ are low-rank representations of object types $X_{h}$ and $X_{l}$   respectively, $S_{hl}$ is a trade-off matrix that provides an additional degree of freedom for the factorization process \cite{Ding:2006}. 

\subsubsection{Diverse Manifold Learning}
The diverse manifold learning is a process to learn the manifold with varied relationships information present in a dataset, i.e., including inter- and intra-relationships in MTRD and including sample-sample and sample-feature relationships in multi-view data utilizing $\{R_{hl}\}$ and $\{W_{h}\}$.

\subsection{Learning Inter- and Intra-manifolds} 
\subsubsection{Motivation}

Existing methods using the manifold learning in MTRD and multi-view data focus on learning the intra-manifold within each data type or within data samples in each data view respectively \cite{Khanh}\cite{Jun}\cite{Zong:2017}. These methods preserve local geometric structure or close distances between the data points of each data type or each view only. However, these methods fail to preserve the geometric structure created by the data points of multiple types or between multiple views. They leave data points of different types sampled on the space without controlling their close distances when mapping to a low-dimensional space. 
This sampling process disregards the relatedness information between the data of different types or views which is valuable for heterogeneous data learning. 

We propose to learn and preserve the inter-manifold for MTRD, based on the assumption that \textit{if two data points, belonging either to the same type or to different data types, are close in original data space (i.e., have high relatedness), their new representations in the new low-dimensional space must also be close}. We formulate the new objective function with the inter-manifold based on this assumption. Since the local geometric information generated from inter-relationship is useful in clustering \cite{Belkin:2003}, we focus on building and maintaining the local geometric structure corresponding to inter-manifold only. 

\subsubsection{Learning Inter-manifold}
We propose to use the $p$ nearest neighbour ($p$NN) graph for constructing a geometric shape of the inter-manifold learned from the high-dimensional original data. The closeness information between data points of two data types $X_h$ and $X_l$ is encoded in $Z_{hl} = \{z_{ij}\}^{n_{h}\times n_{l}}$ that is constructed as, 

\begin{figure}[t]
\centering
\includegraphics[width = 0.66\linewidth]{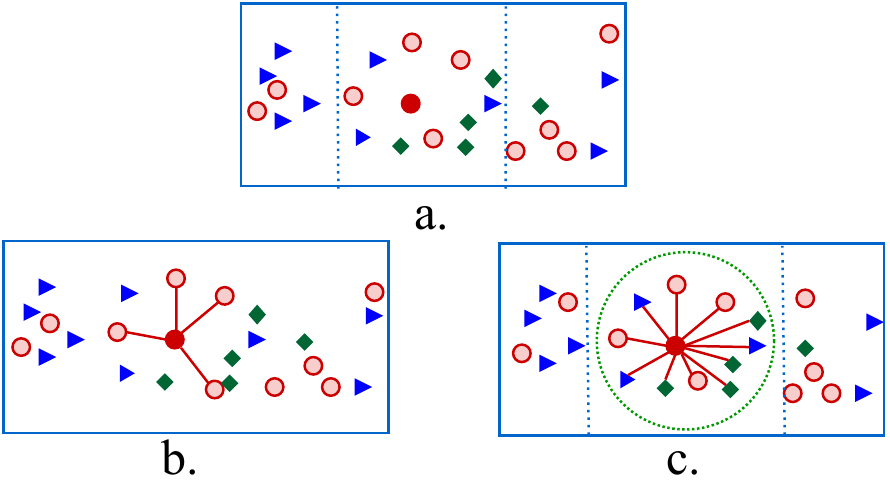}
\caption{Illustration of inter-manifold learning. Consider a MTRD dataset with three data types $X_{1}$, $X_{2}$ and $X_{3}$ lying on manifold in $\textbf{R}^{2}$, Fig. \ref{fig:InterManifold}.a shows the distribution of data in original space, Fig. \ref{fig:InterManifold}.b and Fig. \ref{fig:InterManifold}.c show the data distribution in the mapped low-dimensional space when only the intra-manifold has been used and when both intra- and inter-manifolds have been used, respectively.}	
\label{fig:InterManifold}
\end{figure}

\begin{equation}
z_{ij}=\begin{cases}
r_{ij} \mbox { if } (x_{j} \in  \mathcal{N}_p^{inter}(x_i) \mbox{ or } x_i \in \mathcal{N}_p^{inter}(x_j))\\
0 \mbox{, otherwise}
\end{cases}
\label{eq:Zkl}
\end{equation}
where $\mathcal{N}_p^{inter}(x_i)$ denotes $p$ nearest neighbouring points of $x_i$ of data type $X_h$ . 

$Z_{hl}$ is the weighted adjacency matrix built from a $p$NN graph based on a \textit{scatter of data points} of $X_{h}$ and $X_{l}$. It only stores \textit{high inter-relatedness} between objects of $X_{h}$ and $X_{l}$. In other words, constructing a $p$NN graph for each inter-type relationship matrix corresponds to the process of \textit{discretely approximating} \cite{Gu:2009} the data points of two object types on their intrinsic manifold. It learns distances from a point to all its neighbours from the other types. 

To preserve this closeness information when mapping to a lower-order space, we propose to add the following term: 
\begin{equation}
\min \sum_{h,l=1, h \neq l}^m\sum_{i=1}^{n_{h}}\sum_{j=1}^{n_{l}} \|g_{i}-g_{j}\|^{2}z_{ij} 
\label{eq:Pinter}
\end{equation}
where $\|g_{i}-g_{j}\|^{2}$ is the Euclidean distance estimating closeness between new representations $g_{i}, g_{j}$ projected from $x_{i}, x_{j}$, $x_{i} \in X_h$ and $x_j \in X_l$. 

Note that since $Z_{hl}$ is constructed from $R_{hl}$, one may argue that minimizing Eq. (\ref{eq:Pinter}) is equivalent to minimizing the problem in Eq. (\ref{eq:J1}). However, the objective in two optimization functions is different. Eq. (\ref{eq:J1}) aims to minimize the reconstruction error to learn the low-rank representation while the minimizing problem in Eq. (\ref{eq:Pinter}) ensures preserving the close distances (or the high relatedness) between the data points of different object types in the local geometric structure.

Suppose that $h,l$ are fixed, Eq. (\ref{eq:Pinter}) can be reduced to
\begin{equation}
\min \sum_{i=1}^{n_{h}}\sum_{j=1}^{n_{l}}\norm{g_{i}-g_{j}}^{2}z_{ij} 
\label{eq:P1_1}
\end{equation}
It is equivalent to, 
\begin{equation}
\begin{split}
\min & \big(\sum_{i=1}^{n_{h}}g_{i}g_{i}^{T}\sum_{j=1}^{n_{l}}z_{ij} +  \sum_{j=1}^{n_{l}}g_{j}g_{j}^{T} \sum_{i=1}^{n_{h}}z_{ij}-2\sum_{i=1}^{n_{h}}\sum_{j=1}^{n_{l}}g_{i}g_{j}^{T}z_{ij}\big) 
\end{split}
\label{eq:P1_2}
\end{equation}
\begin{equation}
\begin{split}
\Leftrightarrow \min &\big( Tr(G_{h}^{T}T_{hl}^{r}G_{h})+ Tr(G_{l}^{T}T_{hl}^{c}G_{l}) - 2 Tr(G_{h}^{T}Z_{hl}G_{l})\big) 
\end{split}
\label{eq:P1_3}
\end{equation}
where $T_{hl}^{r}$ is a diagonal matrix of size $n_{h}\times n_{h}$ whose entries are sums of elements in each row of $Z_{hl}$ and  $T_{hl}^{c}$ is a diagonal matrix size $n_{l}\times n_{l}$ whose entries are sums of elements in each column of $Z_{hl}$. More specifically,
\begin{equation}
(T_{hl}^{r})_{ii}=\sum_{j=1}^{n_l}(z_{ij}), 
(T_{hl}^{c})_{jj}=\sum_{i=1}^{n_h}(z_{ij})
\label{eq:T_klrc}
\end{equation}For all $1\leq h \leq m, 1\leq l\leq m, h \neq l$, and from Eq. (\ref{eq:P1_3}), the optimization problem in Eq. (\ref{eq:Pinter}) becomes

\begin{equation}
\begin{split}
\min \sum_{h,l=1,h\neq l}^{m}&\big(Tr(G_{h}^{T}T_{hl}^{r}G_{h})+ Tr(G_{l}^{T}T_{hl}^{c}G_{l}) \\
&-2 Tr(G_{h}^{T}Z_{hl}G_{l})\big)
\end{split}
\label{eq:P1_allkl_1}
\end{equation}
\begin{equation}
\begin{split}
\Leftrightarrow  \min \big(\sum_{h=1}^{m}Tr(G_{h}^{T}T_{h}G_{h}) -2 \sum_{1\leq h < l \leq m} Tr(G_{h}^{T}Q_{hl}G_{l})\big)
\end{split}
\label{eq:P1_final}
\end{equation}
where $Tr(.)$ denotes the trace of a matrix, $T_{h}$ and $Q_{hl}$ are defined as,
\begin{equation}
T_{h} = \sum_{l=1,l\neq h}^{m}(T_{hl}^{r}+T_{lh}^{c})
\label{eq:T_k}
\end{equation}\begin{equation}
Q_{hl} = Z_{hl}+Z_{lh}^{T}
\label{eq:Q_kl}
\end{equation} 

Please refer to appendix A for full transformation from Eq. (\ref{eq:P1_allkl_1}) to Eq. (\ref{eq:P1_final}). 
\begin{figure}[t]
\centering
\includegraphics[width = 0.50\linewidth]{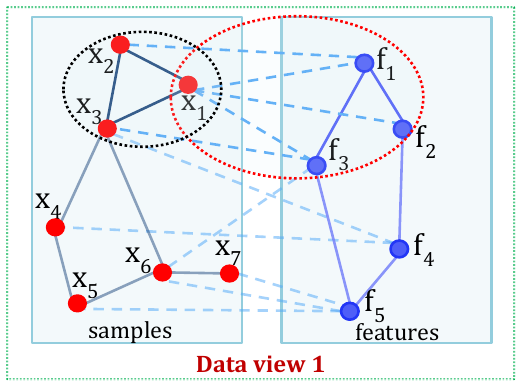}
\caption{Illustration of diverse manifold learning. Distances from a sample ($x_1$) to all its neighbouring samples (red dotted circle) and all its important features (blue dotted circle) are preserved.}	
\label{fig:Sample_feature_Manifold}
\end{figure}

The newly proposed term (Eq. (\ref{eq:P1_final})), named as $P_1$, is to learn \textit{inter-manifold} and makes the proposed method DiMMA distinct from other NMF-based methods \cite{Wang:2011}\cite{Khanh}\cite{Gu:2009}\cite{Jun}. The new component $P_1$ ensures smoothness of the mapping function on objects from different types. Distances between objects (in the neighbourhood area) from different types are well preserved in the newly mapped space. 

Figure \ref{fig:InterManifold} shows how the inter-manifold captures the similarity information between different types in low-dimensional space. Figure \ref{fig:InterManifold}.b illustrates the distribution of data in the mapped space when only the intra-manifold is considered where some highly related data points of different types have not been adequately preserved. Figure \ref{fig:InterManifold}.c illustrates when both intra- and inter-manifold learning are considered and the distances between a point to all its inter- and intra-neighbours have been preserved. This neighbourhood information helps in finding accurate clusters in the newly mapped space. 

When applied to the multi-view data, the proposed inter-manifold will play the role of a sample-feature manifold and will capture similarity information between the sample objects and feature objects that reside on the manifold to ensure the high relatedness between samples and features is preserved in the mapped space. Figure \ref{fig:Sample_feature_Manifold} illustrates a multi-view dataset, in which the sample-feature manifold is learned. Consider the data object $x_1$; by using the sample-feature manifold, all important features of $x_1$ have been preserved and included during the learning process to map the lower-dimensional space.

The component $P_1$ can be integrated into the objective function Eq. (\ref{eq:J1}). Combining the inter-manifold term with the intra-manifold term generates the diverse manifold learning. 

\subsubsection{Combining intra-manifold}
The intra-manifold term on each object type in many recent methods \cite{Wang:2011}\cite{Li:2013} is given as,

\begin{equation}
P_{2} = \min \sum_{h=1}^{m}(Tr(G_{h}^{T}L_{h}G_{h})
\label{eq:P2_final}
\end{equation}
where the graph Laplacian matrix of object type $X_{h}$ is as,
\begin{equation}
L_{h} = D_{h}-W_{h}
\label{eq:Lk}
\end{equation}
$D_{h}$ is the diagonal matrix computed by $(D_{h})_{ii}=\sum_{j}(W_{h})_{ij}$ and $W_{h} = \{w_{ij}\}^{n_{h}\times n_{h}}$ is a weighted adjacency matrix as in Eq. (\ref{eq:intrarelationship}). The term $P_{2}$ (Eq. (\ref{eq:P2_final})) ensures a smoothness of the mapping function on objects within the same type, and ensures that the distances between objects (in the neighbourhood area) from the same type are preserved in a new space. 

We form the diverse manifold $P$ encoding both inter-manifold Eq. (\ref{eq:P1_final}) and intra-manifold Eq. (\ref{eq:P2_final}) linearly as, 
\begin{equation}
P = \delta P1 + \lambda P2, \mbox{ }\delta \geq 0, \lambda \geq 0
\end{equation}

The diverse manifold term $P$ is incorporated into the objective function Eq. (\ref{eq:J1}) to form the proposed and novel MTRD learning objective function as,

\begin{equation}
\begin{split}
\min &\sum_{1\leq h<l\leq m}\big(\norm{R_{hl}-G_{h}S_{hl}G_{l}^{T}}_{F}^{2}-2\delta Tr(G_{h}^{T}Q_{hl}G_{l}) \big) \\
+\lambda &\sum_{h=1}^{m} Tr(G_{h}^{T}L_{h}G_{h})  + \delta \sum_{h=1}^{m} Tr(G_{h}^{T}T_{h}G_{h})  \\
& \text{s.t., }G_{h}\geq 0, G_{l} \geq 0, G_{h}1_{c}=1_{n_{h}},G_{l}1_{c}=1_{n_{l}},\\
&\forall 1\leq h < l \leq m
\end{split}
\label{eq:preJ2}
\end{equation}
where $1_{c},1_{n_{h}}, 1_{n_{l}}$ are all-ones column vectors with sizes $c, n_{h}, n_{l}$ respectively, with every element is $1$. The $l1$-normalization on $G_h$ and $G_l$ makes them more meaningful in ranking the values \cite{Jun}. $\lambda$ and $\delta$ are intra regularization and inter regularization parameters, respectively. 
If we set 
\begin{equation}
Q_{h} = \lambda L_{h}+ \delta T_{h}
\label{eq:QkQl}
\end{equation}

Eq. (\ref{eq:preJ2}) will become as follows:

\begin{equation}
\begin{split}
& J_2 = \\
&\min \sum_{1\leq h<l\leq m}\big(\norm{R_{hl}-G_{h}S_{hl}G_{l}^{T}}_{F}^{2} - 2\delta Tr(G_{h}^{T}Q_{hl}G_{l}) \big) \\+
&\sum_{h=1}^{m} Tr(G_{h}^{T}Q_{h}G_{h}) 
\text{ s.t., }G_{h}\geq 0, G_{l} \geq 0, \\
&G_{h}1_{c}=1_{n_{h}}, G_{l}1_{c}=1_{n_{l}},\forall 1\leq h < l \leq m
\end{split}
\label{eq:J2}
\end{equation}

The new objective function has three terms that assist DiMMA to simultaneously learn the low dimensional representations for all object types (the first term) that respect the intra and inter-manifolds (second and third terms). Two parameters $\lambda$ and $\delta$ can be adjusted. In particular, larger  $\lambda$ and $\delta$ values should be set if the dataset is known to be lying on manifolds. The $m$ and $m(m-1)/2$ different values for $\lambda$ and $\delta$ can be set to allow these parameters to act as weights of each object type and each inter-relationship type. In this paper, we give equal weight to each term to treat these object types and these relationships equally. 

Next, we summarize the important properties of the proposed diverse manifold.

\textbf{Lemma 1}

\textit{Minimizing the diverse manifold term P will ensure close distances between an object point to all its inter- and intra-neighbourhood points preserved. }

The neighbourhood constraint is defined as follows. A pair of two points $(x_{i},x_{j})$ is called to satisfy the neighbourhood constraint if they satisfy the inter- or intra-neighbourhood constraint. A pair of two points $(x_{i},x_{j})$ belonging to two data types $X_{h},X_{l}$ is called to satisfy the inter-neighbourhood constraint if $x_{j}\in \mathcal{N}^{p}_{inter}(x_{i})$ or $x_{i}\in \mathcal{N}^{p}_{inter}(x_{j})$ where $\mathcal{N}^{p}_{inter}(x_{i})$ denotes the $p$ nearest inter-type neighbour points of $x_{i}$.   A pair of two points $(x_{i},x_{j})$ from the same data type $X_{h}$, is called to satisfy the intra-neighbourhood constraint if $x_{i}\in \mathcal{N}^{p}_{intra}(x_{j})$ or $x_{j}\in \mathcal{N}^{p}_{intra}(x_{i})$, where $\mathcal{N}^{p}_{intra}(x_{i})$ denotes the $p$ nearest neighbour points of $x_{i}$.

\textbf{Proof: }
Minimizing P ensures that the close distance between $x_i$ and $x_j$ will be preserved when $x_i$, $x_j$ are projected to a lower-dimensional space.

Considering an arbitrary pair of two points $(x_{i}, x_{j})$ that satisfies the neighbourhood constraint, we have

(1) points $(x_{i}, x_{j})$ with a high relatedness $z_{ij}$, or a close distance value $d(x_{i}, x_{j})$ in the input space.

(2) $z_{ij}>0$ from the definition of $z_{ij}$ in Eq. (\ref{eq:Zkl}).  

Minimizing $\norm{g_{i}-g_{j}}^{2}z_{ij}$, s.t. $z_{ij}>0$ leads to minimizing $\norm{g_{i}-g_{j}}^{2}$ or equivalently keeping distance $d(g_{i}, g_{j})$ between $g_i$ and $g_j$ close.

From (1) and (2), we conclude that the close distance between $x_i, x_j$ in the original space has been preserved on the newly mapped points $g_i, g_j$ in the new lower-dimensional space through minimizing P. This fact is true on every pair of points that holds the neighbourhood constraint. 

The value of $z_{ij}$ is set to $0$ when $(x_{i}, x_{j})$ is unable to satisfy the neighbourhood constraint. It shows that points reside far away from each other. It ensures that the minimizing process only preserves the distances between data points that are close enough in the input space. It maintains the high relatedness between similar objects from the same type or the different types while it ignores others.

\textbf{Lemma 2}

\textit{Minimizing the diverse manifold term P results in a ranking of features represented by objects of different types.  }

\textbf{Proof: }

The objective function in Eq. (\ref{eq:J2}) includes the following inter-manifold term,
\begin{equation}
\begin{split}
&\min \sum_{1\leq h<l\leq m}(-Tr(G_{h}^{T}Q_{hl}G_{l}))\\
\Leftrightarrow &\max \sum_{1\leq h<l\leq m}Tr(G_{h}^{T}Q_{hl}G_{l})
\end{split}
\label{eq:dis1}
\end{equation}

Eq. (\ref{eq:dis1}) can be rewritten and represented as: 
\begin{equation}
\sum_{1\leq h<l\leq m}Tr(G_{h}^{T}Q_{hl}G_{l})=\sum_{h=1}^{m-1}\sum_{i=1}^{n_{h}}\sum_{l=h+1}^{m}q_{i}G_{l}g_{i}^{'}
\label{eq:dis3}
\end{equation}
where $g_{i}=[g_{i1}, g_{i2}, ..., g_{ic}]$ and $q_{i}=[q_{i1}, q_{i2}, ..., q_{in_{l}}]$

Consider a particular object $x_{i}\in X_{h}$, using Eq. (\ref{eq:dis3}), we have,
\begin{equation}
\sum_{l=h+1}^{m}q_{i}G_{l}g_{i}^{'}=\sum_{t=1}^{c}g_{it}\sum_{l=h+1}^{m}\sum_{j=1}^{n_{l}}q_{ij}g_{jt}
\label{eq:dis4}
\end{equation}

Let $Y=\sum_{l=h+1}^{m}\sum_{j=1}^{n_{l}}q_{ij}g_{jt}$. 
Maximizing Eq. (\ref{eq:dis4}) forces the learning process to assign a large $g_{it}$ value to $x_{i}$ if its related $Y$ is large. This is because $l1$-norm on $G_{h}$ allows producing many coefficients with zeros or very small values with few large values \cite{l1norm}. Consequently, $Y$ can be seen as a ranking term for an object type in Eq. (\ref{eq:J2}). Considering features of objects of different types, assigning a higher value to $g_{it}$ when $Y$ has a promising high value implies that the most important features were included during the clustering process. Hence, it can be stated that DiMMA supports feature selection while clustering. 

\subsection{Algorithmic Solution to the DiMMA function}
\subsubsection{The DiMMA Algorithm}
We provide a solution to the proposed objective function in Eq. (\ref{eq:J2}) with respect to  $S_{hl}, G_{h}, G_{l}$. We will separately update each variable while fixing others as constant until convergence \cite{Wang:2011}\cite{Gu:2009}\cite{Li:2013} and introduce the iterative Algorithm 1 to simultaneously group several type objects into clusters. 

\textbf{1.	Solving $S_{hl}$:} When fixing $h,l$ and thus fixing $G_{h}, G_{l}$,  Eq. (\ref{eq:J2}) with respect to $S_{hl}$ is reduced to minimizing 
\begin{equation}
J_{S_{hl}} = \norm{R_{hl}-G_{h}S_{hl}G_{l}^{T}}_{F}^{2}
\label{eq:J_Skl}
\end{equation}
\begin{equation}
\partial J_{S_{hl}}/\partial S_{hl} = -2R_{hl}G_{h}^{T}G_{l} + 2S_{hl}G_{l}^{T}G_{l}G_{h}^{T}G_{h}
\end{equation}
\begin{equation}
\Leftrightarrow 2R_{hl}G_{h}^{T}G_{l} = 2S_{hl}G_{l}^{T}G_{l}G_{h}^{T}G_{h}
\end{equation}

Then we have the update rule for $S_{hl}$
\begin{equation}
S_{hl} = (G_{h}^{T}G_{h})^{-1}G_{h}^{T}R_{hl}G_{l}(G_{l}^{T}G_{l})^{-1}
\label{eq:S}
\end{equation}

\textbf{2.	Solving $G_{h}, G_{l}$:} Solving $G_{h}, G_{l}, 1\leq h<l \leq m$ is obviously equivalent to optimize all $G_h$ where $1 \leq h \leq m$.   
When fixing $h$, fixing $S_{hl}, G_l, h<l\leq m$, fixing $S_{lh}, G_l, 1\leq l<h$, we can rewrite the objective function in Eq. (\ref{eq:J2}) as follows:
\begin{equation}
\begin{split}
&J_{G_{h}} =  Tr(G_{h}^{T}Q_{h}G_{h}) + \\
&\sum_{h<l\leq m}\Big(\norm{R_{hl}-G_{h}S_{hl}G_{l}^{T}}_{F}^{2}-\delta Tr(G_{h}^{T}Q_{hl}G_{l}) \Big)\\
&+ \sum_{1\leq l<h} \Big(\norm{R_{lh}-G_{l}S_{lh}G_{h}^{T}}_{F}^{2}-\delta Tr(G_{l}^{T}Q_{lh}G_{h}) \Big)\\
\end{split}
\label{eq:J_Gh}
\end{equation}

By taking the derivative of $J_{G_{h}}$ on $G_{h}$ we have:
\begin{equation}
\begin{split}
& \partial J_{G_{h}}/\partial G_{h}= 2G_{h}Q_{h}  \\
&+ \sum_{h<l \leq m} \Big( -2R_{hl}G_{l}S_{hl}^{T} + 2G_{h}S_{hl}G_{l}^{T}G_{l}S_{hl}^{T}\Big)-\delta Q_{hl}G_{l}\Big)\\
&+ \sum_{1\leq l<h} \Big(-2R_{lh}^{T} G_{l}S_{lh} + 2G_{h}S_{lh}^{T}G_{l}^{T}G_{l}S_{lh}- \delta Q_{lh}^{T}G_l\Big)
\end{split}
\end{equation}
\begin{equation}
\begin{split}
A_h & = \sum_{h<l \leq m} ( -R_{hl}G_{l}S_{hl}^{T}-1/2\delta Q_{hl}G_l) \\
& + \sum_{1\leq l<h} (-R_{lh}^{T} G_{l}S_{lh}-1/2\delta Q_{lh}^{T}G_{l})
\end{split}
\end{equation}
\begin{equation}
B_h = \sum_{h<l \leq m}(S_{hl}G_{l}^{T}G_{l}S_{hl}^{T})+ \sum_{1\leq l<h}(S_{lh}^{T}G_{l}^{T}G_{l}S_{lh})
\end{equation}

By introducing the Lagrangian multipler matrix $\Lambda$ and setting the partial derivative of $G_{h}$ to $0$, we obtain 
\begin{equation}
\Lambda = 2\lambda Q_hG_{h} - 2A_{h} + 2G_{h}B_{h}
\end{equation}

Since the Karush-Kuhn-Tucker (KKT) condition \cite{KKT:2004} for the non-negative constraint on $G_{h}$ gives $(\Lambda)_{ij}(G_{h})_{ij} = 0$, we have the following update rule for $G_{h}$, 
\begin{equation}
(G_{h})_{ij} = (G_{h})_{ij}\left[ \dfrac{(Q_{h}^{-}G_{h} + A_{h}^{+} + G_{h}B_{h}^{-})_{ij}}{(Q_{h}^{+}G_{h} + A_{h}^{-} + G_{h}B_{h}^{+})_{ij}} \right]^{1/2} 
\label{eq:Gh}
\end{equation}

where $(Q_{h}^{+})_{ij}=(\abs{(Q_{h})_{ij}}+(Q_{h})_{ij})/2$, $(Q_{h}^{-})_{ij}=(\abs{(Q_{h})_{ij}}-(Q_{h})_{ij})/2$ and similarly to $A_{h}^{+/-}$, $B_{h}^{+/-}$ \cite{Ding:2010}.

\RestyleAlgo{boxruled}
\begin{algorithm}[t]
\SetKwInOut{Input}{Input}\SetKwInOut{Output}{Output}
\Input{Inter relationship matrices $\{R_{hl}\}_{1\leq h<l \leq m}^{n_{h}\times n_{l}}$, clusters number $c$, intra graph parameters $\lambda$, inter graph parameter $\delta$, intra and inter neighbourhood size $k,p$.}
\Output{Cluster labels.}
\BlankLine
Initialize non-negative matrices $\{G_{h}\}_{1 \leq h \leq m}$ by K-means, using inter relationship matrices $\{R_{hl}\}_{1\leq h<l \leq m}^{n_{h}\times n_{l}}$ and clusters number $c$ as input.\

\textbf{1.} \For {each $h, 1 \leq h \leq m$} {
	construct Intra affinity matrices $\{W_{h}\}_{1\leq h \leq m}$ as in Eq. (\ref{eq:intrarelationship})
	
	compute the Laplacian matrices $L_{h}$ as in Eq. (\ref{eq:Lk}).
}

\textbf{2.} \For {each $(h,l), 1\leq h \neq l \leq m$} {
	construct the affinity matrix $Z_{hl}$ as in Eq. (\ref{eq:Zkl}).
	
	construct $T_{hl}^r$ and $T_{hl}^c$ as in Eq. (\ref{eq:T_klrc})
}
\textbf{3. }\For {each $h, 1 \leq h \leq m$} {
	construct $T_{h}$ as in Eq. (\ref{eq:T_k})
	
	construct $Q_{h}$ as in Eq. (\ref{eq:QkQl})
}
\textbf{4. }\For {each $(h,l), 1\leq h<l \leq m$} {
	construct $Q_{hl}$ as in Eq. (\ref{eq:Q_kl})
}
\textbf{5. }\Repeat{convergence}{
	
	\For {each $(h,l), 1\leq h<l \leq m$}{update $S_{hl}$ as in Eq. (\ref{eq:S})
	}
	
	\For {each $h, 1 \leq h \leq m$}{ 
		Update $G_{h}$ as in Eq. (\ref{eq:Gh})
		
		Normalize $G_{h}$		
	}
	
}
\textbf{6.} Transform $\{G_{h}\}_{1\leq h \leq m}$ into cluster indicator matrices by K-means.
\caption{Learning Diverse Manifold for Multi-Aspect Data (DiMMA)\label{alg}}
\end{algorithm}

Algorithm $1$ summarizes the processing of DiMMA. In this paper, we initialize the low-rank matrices $G_{h}$ based on the solution running the partitional algorithm K-means on the inter relationship matrices $\{R_{hl}\}$. Note that a different initializing technique such as random initialization or another clustering method can also be used. However, the use of an initialization method does not have a significant impact on the final solution and the convergence. By iteratively updating $G_{h}$ as in Eq. (\ref{eq:Gh}), DiMMA includes not only intra-type graphs but also includes inter-type graphs and returns the best result after convergence. The optimizing steps are the same as an NMF-based method incorporating intra-manifold learning, with the addition of the process of constructing the \textit{p}-nearest neighbour (pNN) graphs, $\{Z_{hl}\}_{1\leq h\neq l \leq m}$. Using nearest neighbours, $Z_{hl}$ can be built directly from $R_{hl}$ without the need of computing distances. Consequently, this step significantly improves the overall complexity of the algorithm. Similarly, in the last step, K-means or any other clustering algorithm can be applied to the newly learned representation to obtain cluster labels. 
\begin{table*}
	\centering
	\caption{Characteristic of the datasets}
	\label{tab:dataset}       
	\begin{tabular}{|l|r|r|r|r|r|r|}
		\hline
		\textbf{Properties}&\textbf{D1 (MLR-1)}		&\textbf{D2 (MLR-2)}	&\textbf{D3 (R-Top)}		&\textbf{D4 (R-MinMax)}		&\textbf{D5 (Movie)}		&\textbf{D6 (Caltech)}	\\
		
		\hline
		$\#$ Samples/ $\#$ Classes/ $\#$ Object types	& 8,400 / 6/ 5 				& 3,600/ 6/ 5			& 4,000/ 10/ 3			&  1,413 / 25/ 3				& 617/ 17/ 3				& 2,386 / 20/ 6 		\\ \hline
		$\#$ English terms/ Gabor 	& 5,000 			& 4,000		& - 		& -					& - 				& 48 			\\ \hline
		$\#$ French terms/ Centrist 	& 5,000 			& 4,000 	& -				& - 				& - 				&254 			\\ \hline
		$\#$ German terms/ HOG 	& 5,000  			& 4,000 	& - 			& - 				& - 				& 1,984 			\\ \hline
		$\#$ Italian terms/ GIST 	& 5,000  			& 4,000 	& - 			& - 				& - 				& 512 			\\ \hline
		$\#$ Terms/ LBP/ Actors 	& -  				& - 		& 6,000 		& 2,921 			& 1,398 				& 928 			\\ \hline
		$\#$ Concepts/ Keywords	& -  				& - 		& 6,000 		& 2,437				& 1,878 				& - 			\\ \hline
	\end{tabular}
	\vspace{1em}
	\begin{threeparttable}[b]
		\caption{NMI of each dataset and method (Percent)}
		\label{tab:NMI}       
		\begin{tabular}{|l|r|r|r|r|r|r|r|}
			\hline
			\textbf{Methods}&\textbf{D1 (MLR-1)}		&\textbf{D2 (MLR-2)}	&\textbf{D3 (R-Top)}		&\textbf{D4 (R-MinMax)}		&\textbf{D5 (Movie)}		&\textbf{D6 (Caltech)}	&\textbf{Average}\\\hline
			
			NMF 			& 26.41 			& 22.82 			& 30.02 		& 55.16 			& 15.80 			& 47.59		&32.97		\\ \hline
			DRCC 			& 15.01 			& 14.14 			& 46.04 		& 63.45 			& 22.41 			& 55.53 	&36.10		\\ \hline
			STNMF 			& 15.50 			& 19.42 			& 41.10			& 67.28 			& 27.59 			& 38.94 	&34.97		\\ \hline
			RHCHME 			& 24.32 			& 23.17 			& 48.90 		& 69.28 			& 31.02 			& 34.24 	&38.49	\\ \hline
			DRCC-Extended 	& 22.43 			& 20.60 			& 46.27 		& 71.64 			& 29.57 			& 53.93 	&40.74	\\ \hline
			MMNMF 			& 30.86 			& 30.59 			& 44.11 		& 60.50 			& 23.11 			& 42.98 		&38.69	\\ \hline
			\textbf{DiMMA} & \textbf{34.96} 	& \textbf{30.76} & \textbf{50.38} 	& \textbf{73.06} & \textbf{33.87} 		& \textbf{56.48} &\textbf{46.59}\\\hline
		\end{tabular}

	\end{threeparttable}
	\vspace{1em}
	\caption{Accuracy of each dataset and method (Percent)}
	\label{tab:ACC}       
	\begin{tabular}{|l|r|r|r|r|r|r|r|}
		\hline
		\textbf{Methods}&\textbf{D1 (MLR-1)}		&\textbf{D2 (MLR-2)}	&\textbf{D3 (R-Top)}		&\textbf{D4 (R-MinMax)}		&\textbf{D5 (Movie)}		&\textbf{D6 (Caltech)}	&\textbf{Average}\\\hline
		NMF 			& 44.19 			& 39.94 			& 27.55 		& 46.21 			& 17.02 			& 32.52 	&34.57		\\ \hline
		DRCC 			& 31.80 			& 24.58 			& 49.85 		& 53.22 			& 22.53 			& 46.48 	&38.08		\\ \hline
		STNMF 			& 36.67 			& 24.44			 	& 50.30			& 51.38 			& 26.26 			& 31.68 	&36.79		\\ \hline
		RHCHME 			& 44.07 			& 42.33 			& 53.12	 		& 56.62 			& 30.31 			& 30.22 	&42.79		\\ \hline
		DRCC-Extended 	& 43.19 			& 39.75 			& 47.25 		& 59.45 			& 26.90 			& 46.40 	&43.82		\\ \hline
		MMNMF 			& 45.23 			& 44.19			& \textbf{52.25}		& 45.65				& 20.58			& 35.75 	&42.28		\\\hline
		\textbf{DiMMA} & \textbf{49.38} 	& \textbf{46.50} 	& 51.10 & \textbf{60.16} 	& \textbf{31.44} 		& \textbf{54.23} &	\textbf{48.80}\\\hline
	\end{tabular}

\end{table*}
\subsubsection{Convergence of the Algorithm}
We prove the convergence of the DiMMA Algorithm through update rules of $S_{hl}, G_{h}$ as shown in Eqs. (\ref{eq:S}) and (\ref{eq:Gh}) respectively by using the auxiliary function approach \cite{NMF2001}.

\textit{$Z(h,h')$ is an auxiliary function for $F(h)$ if conditions 
$Z(h,h') \geq F(h)$ and $Z(h,h) = F(h)$ are satisfied \cite{NMF2001}.} 

\textbf{Lemma 3: } 
\textit{If $Z$ is an auxiliary function for $F$, then $F$ is non-increasing under the update \cite{NMF2001}}
$$h^{(t+1)} = arg \min_{h} Z(h,h^{(t)})$$

\textbf{Proof:} $F(h^{(t+1)}\leq Z(h^{(t+1)}, h^{(t)}) \leq Z(h^{(t)}, h^{(t)}) = F(h^{(t)})$.

\textbf{Lemma 4: } \textit{For any non-negative matrices $A \in \Re^{n \times n}, B \in \Re^{k \times k}, S \in \Re^{n \times k}, S' \in \Re^{n \times k}$ where $A$ and $B$ are symmetric, the following inequality holds \cite{Ding:2010}}

$$\sum_{i=1}^n \sum_{p=1}^{k} \dfrac{(AS'B)_{ip}S_{ip}^{2}}{S'_{ip}} \geq Tr(S^{T}ASB)$$

Next, we present two theorems that show the convergence of the DiMMA algorithm relating to the used auxiliary function. 

\textbf{Theorem 1:} 
\textit{Let} 
\begin{equation}
L(G_h) = Tr(G_h^{T}Q_hG_h) - A_hG_h^{T} + G_hB_hG_h^{T})
\label{eq:L(G)}
\end{equation}

\textit{then the following function }

$Z(G_h,G_h') = \sum_{ij}\dfrac{(Q_h^{+}G_h')_{ij}(G_h)_{ij}^2}{(G_h')_{ij}} -$ $\sum_{ijk}(Q_h^{-})_{jk}(G_h')_{ji}(G_h')_{ki}(1+log\dfrac{(G_h)_{ji}(G_h)_{ki}}{(G_h')_{ji}(G_h')_{ki}})$

$- 2\sum_{ij}(A_h^{+})_{ij}(G_h')_{ij}(1+log\frac{(G_h)_{ij}}{(G_h')_{ij}}) +2\sum_{ij}(A_h^{-})_{ij}\dfrac{(G_h^{2})_{ij}+(G_h'^{2})_{ij}}{2G'_{ij}}	$
$+ \sum_{ij}\dfrac{(G_h'B_h^{+})_{ij}(G_h^{2})_{ij}}{(G_h')_{ij}} $
$-\sum_{ijk}(B_h^{-})_{jk}(G_h')_{ij}(G_h')_{ik}(1+log\dfrac{(G_h)_{ij}(G_h)_{ik}}{(G_h')_{ij}(G_h')_{ik}})$

\textit{is an auxiliary function for $L(G_h)$. It is a convex function in $G_h$ and its global minimum is }
$$(G_{h})_{ij} = (G_{h})_{ij}\left[ \dfrac{(Q_{h}^{-}G_{h} + A_{h}^{+} + G_{h}B_{h}^{-})_{ij}}{(Q_{h}^{+}G_{h} + A_{h}^{-} + G_{h}B_{h}^{+})_{ij}} \right]^{1/2}$$

\textbf{Proof:} See Appendix B. 

\textbf{Theorem 2: }
\textit{Updating ${G_h}, 1\leq h\leq m$ using the update rule as in Eq. (\ref{eq:Gh}) will monotonically decrease the objective function as in Eq. (\ref{eq:J2})}

\textbf{Proof:} According to Lemma 3 and Theorem 1, we have 
$$L(G_{h}^{0}) = Z(G_{h}^{0},G_{h}^{0}) \geq Z(G_{h}^{1},G_{h}^{0}) \geq L(G_h^1) \geq ...$$ 

Therefore $L(G_h)$ is monotonically decreasing. 

Theorems 1 and 2 guarantee the convergence of DiMMA regarding $G_h$. The sub-problem in Eq. (\ref{eq:J_Skl}) is a convex problem regarding $S_{hl}$ with the global minima in Eq. (\ref{eq:S}). The correctness of the algorithm is also guaranteed as the update rules in Eq. (\ref{eq:Gh}) satisfies the Karush-Kuhn-Tucker optimal condition \cite{KKT:2004}. Thus, we conclude that DiMMA monotonically decreases the objective function in Eq. (\ref{eq:J2}) and converges to an optimal solution. 

\subsubsection{Computational Complexity Analysis}
The computational complexity of DiMMA includes three main components: intra-manifold learning; inter-manifold learning and applying multiplicative updating. 

The cost of learning intra-manifold is $O(n_h^{2}km)$. 

The cost of learning inter-manifold, steps $2-4$ in the DiMMA algorithm, is $O(n_hn_lm(m-1))$.

The cost of applying multiplicative updating includes the cost for updating $\{S_{hl}\}_{1\leq h < l \leq m}$, $O(n_hn_lc\dfrac{m(m-1)}{2})$ and updating $\{G_h\}_{1 \leq h \leq m}$, $O(n_h^2cm)$.  

The overall time complexity of DiMMA becomes: $$O(n_h^{2}km + n_hn_lm(m-1) + n_hn_lc\dfrac{m(m-1)}{2}))$$

where $m$ is the number of data types, $k$ is the number of nearest neighbours, $c$ is the number of clusters, and $n_h, n_l$ are the number of objects of object types $X_h, X_l$, respectively. 

The size of every object type is about the same. The values of $m$ and $c$ are comparatively much smaller than the object type size. Hence, the computational complexity of DiMMA remains quadratic. This is similar to existing MTRD algorithms and the additional similarity computation will not incur extra cost with the benefit of improving the accuracy. We show this in Section 4 with extensive experiments.

\section{Empirical analysis}
\label{sec:experiment}
\subsection{Datasets}
We use several real-world datasets to evaluate the performance of DiMMA (Table \ref{tab:dataset}). MLR-1 (D1) and MLR-2 (D2) were created from the popular Reuters RCV1/ RCV2 Multilingual dataset \cite{NIPS2009_3690}. For the MTRD setting, object types include original English documents, English terms, French terms, German terms and Italian terms. The inter-type relationships are based on the appearances of each translated language terms in documents. Each intra-type relationship is generated based on the co-occurrence of corresponding terms in the same language documents. R-Top (D3) and R-MinMax (D4) were selected from Reuters-21578, a well-known text dataset. D3 contains the 10 largest classes in Reuters-21578 and D4 includes 25 classes of the original dataset with at least 20, and at most 200 documents per class. To create MTRD, we used external knowledge i.e., Wikipedia and followed the process as in \cite{Jun}\cite{Jing:2011} to generate the third data type concept (along with document and term data types). D3 and D4 have three inter-relationships: documents-terms; documents-concepts and terms-concepts. The intra-affinity matrix on each object type is created by following the steps in \cite{Wang:2011}\cite{Cai:2011}\cite{Gu:2009}. Movie (D5) and image Caltech-101 (D6) \cite{Caltech-101} datasets have been popularly used in clustering evaluation \cite{KhanhICDE2020}\cite{Co-regularized:2011}\cite{Hussain:2014}. D5 contains the set of movies represented by two different views that are movies-actors and movies-keywords. 
D6, a subset of the image Caltech-101 \cite{Caltech-101}, contains 20 classes with five different views: Gabor, Centrist, HOG, GIST, and LBP.   

\begin{table*}[t]
	\centering
	\vspace{1em}
	\caption{A comparison with Deep Learning-based Low-rank Representation method (Performance on 20\% Test Data)}
	\label{tab:DLcompare}       
	\begin{tabular}{|l|r|r|r|r|r|r|r|r|r|r|r|r|}
		\hline
		\multirow{2}{*}{\textbf{Methods}} & \multicolumn{2}{ |r| }{\textbf{D1 (MLR-1)}} & \multicolumn{2}{ |r| }{\textbf{D2 (MLR-2)}} & \multicolumn{2}{ |r| }{\textbf{D3 (R-Top)}} & \multicolumn{2}{ |r| }{\textbf{D4 (R-MinMax)}} & \multicolumn{2}{ |r| }{\textbf{D5 (Movie)}} & \multicolumn{2}{ |r| }{\textbf{D6 (Caltech)}}\\
		
		\cline{2-13}
		&\textbf{NMI}&\textbf{AC}		&\textbf{NMI}&\textbf{AC}	&\textbf{NMI}&\textbf{AC}		&\textbf{NMI}&\textbf{AC}	&\textbf{NMI}&\textbf{AC}		&\textbf{NMI}&\textbf{AC}	\\\hline
		DCCA 			& 29.15 			& 47.68 			& 30.64 		& 42.22 			& 68.44 			& 53.36 	&45.18&47.25&34.97&26.83&34.97&26.83		\\ \hline
		DeepMVC 		& 2.10 			& 24.52 			& 7.30 		& 26.39 			& 12.82 			& 26.63 	&43.24&33.22&40.35&26.02&42.35&34.59		\\ \hline
		DiMMA-ExInter & 71.60 		& 95.05 			& 63.38 		& 91.71 			& 75.98				& 61.63 	&62.64&58.09&39.09&34.21&39.09&34.21		\\ \hline
		DiMMA 			& \textbf{75.32} 			& \textbf{95.64}		 	& \textbf{77.42}			& \textbf{96.31}				& \textbf{78.58} 			& \textbf{66.28} 	&\textbf{67.85}&\textbf{65.15}&\textbf{49.82}&\textbf{36.84}&\textbf{49.82}&\textbf{36.84}		\\ \hline
		
	\end{tabular}
\end{table*}

\begin{table*}
	\centering
	\vspace{1em}
	\caption{Feature Selection Performance (Cohesiveness of Cluster Solution) of DiMMA}
	\label{tab:cohes}       
	\begin{tabular}{|l|r|r|r|r|r|r|r|r|r|r|}
		\hline
		\textbf{Methods}&\textbf{D1 (MLR-1)}&\textbf{D2 (MLR-2)}&\textbf{D3 (R-Top)}&\textbf{D4 (R-MinMax)}		&\textbf{D5 (Movie)}	&\textbf{D6 (Caltech)}	\\\hline
		STNMF 			&0.0043&0.0094&0.0676& 0.5538	& 0.5165 				& 0.3229 			\\ \hline
		RHCHME 			&0.0046&\textbf{0.0099}&0.0705& 0.5987	& 0.5394			 	& 0.3417		\\ \hline
		DRCC-Extended 	&0.0045&0.0095&0.0702& 0.5964	& 0.5305 				& 0.3416		\\ \hline
		\textbf{DiMMA}  &\textbf{0.0046}&0.0097&\textbf{0.0706}& \textbf{0.5999}& \textbf{0.5428} 	& \textbf{0.3421} \\\hline
	\end{tabular}
\end{table*}

\begin{table}
	\vspace{1em}
	\caption{Running time of each dataset and method (thousand seconds)}
	\label{tab:time}      
	\begin{tabular}{|l|r|r|r|r|r|r|r|}
		\hline
		\textbf{Methods}&\textbf{D1}		&\textbf{D2}	&\textbf{D3}		&\textbf{D4}		&\textbf{D5}		&\textbf{D6}	\\\hline
		NMF 			& 0.79 				& 0.26 				& 1.06 			& 0.05 			& 0.015 	& 0.07 	\\ \hline
		DRCC 			& 0.81 				& 0.26 				& 0.46 			& 0.05 			& 0.015 	& 0.06  	\\ \hline
		STNMF 			& 9.46 				& 5.00 				& 1.63			& 0.26 			& 0.060 	& 8.16 		\\ \hline
		RHCHME 			& 127.59 			& 36.83			 	& 1.63			& 0.26 			& 0.065 	& 1.72 		\\ \hline
		DRCC-Extended 	& 3.43 				& 1.35 				& 1.06 			& 0.16 			& 0.059 	& 0.12 		\\ \hline
		MMNMF 			& 2.55				& 0.49				& 0.41			& 0.07 			& 0.020		& 0.63	\\ \hline
		\textbf{DiMMA} 	& 2.55 				& 1.39 				& 1.06			& 0.16 			& 0.057 	& 0.12		\\\hline
	\end{tabular}
\end{table}

\subsection{Benchmarking methods}

\label{sec:benchmarkingmethods}
DiMMA is compared with traditional NMF \cite{NMF2001}, DRCC co-clustering \cite{Gu:2009} and its variation applicable to multi-type, the state-of-the-art NMF based MTRD clustering methods STNMF \cite{Wang:2011} and RHCHME \cite{Jun} and multi-view method MMNMF \cite{Zong:2017}. Traditional NMF is used to ascertain that MTRD methods should give better performance as they use more information. DRCC (Dual Regularized Co-Clustering) \cite{Gu:2009} is a co-clustering method that considers sampling on the manifold for both data objects and features. It considers intra-manifold only. To enable comparison between DRCC and DiMMA, we extended DRCC by constructing a graph Laplacian $L_h$ for each object type $X_h$ in the dataset. STNMF (Symmetric Non-negative Matrix Tri-Factorization) \cite{Wang:2011} is a leading MTRD clustering method and comes closest to DiMMA. It considers the intrinsic manifold on every object type by constructing a Laplacian graph corresponding to the data type. Different inter-type relationships of different object types are encoded in a symmetric matrix. RHCHME (Robust High-order Co-Clustering via Heterogeneous Manifold Ensemble) \cite{Jun} extends the STNMF framework with a focus on learning complete and accurate intra-relationship information by considering data lying on manifolds and data belonging to subspaces. The multi-view method, MMNMF \cite{Zong:2017}, learns the consensus low-rank representation of all views data relying on learning and preserving the consensus manifold generated from different views.

Additionally, we benchmarked DiMMA with deep-model-based methods including a leading deep learning method DCCA \cite{DCCA:2013} and the deep matrix factorization (DeepMVC)  \cite{DeepMVC}. We present this comparison in a separate section.

\subsection{Experimental settings}
All benchmarking methods and DiMMA produce a low-rank representation. We use a centroid-based clustering method, K-means, to obtain cluster labels from the learned low-rank representations for all methods. The number of clusters on each dataset $K$ or $c$ is set equal to the number of actual classes in datasets. To use NMF, we choose the best view of each dataset, i.e., between documents and English terms on datasets D1 and D2; between documents and terms on D3 and D4; between movies and actors on D5; and using GIST representation on D6. The Co-clustering method DRCC uses the same information as NMF, but co-clusters both objects and features simultaneously. 

We utilize the two widely used measurement criteria, clustering accuracy (AC), the percentage of correctly obtained labels and Normalized Mutual Information (NMI) \cite{NMI1}. We also report the computational time consumed by all methods. Average results are produced after the 5-fold runs of each experiment. All methods are implemented in Matlab.

\subsection{Accuracy of Clustering solution}

Results in Tables \ref{tab:NMI} and \ref{tab:ACC} show that DiMMA outperforms all benchmarking methods on all datasets (except MMNMF achieves marginally better accuracy on D3). The reason behind DiMMA's superiority is its capability of utilizing the local structures of objects from the same type and different types based on the proposed diverse manifold learning. This asserts the effectiveness of incorporating the manifold learning on both intra and inter-type relationships into an NMF-based method. DiMMA performs exceptionally better as compared to traditional NMF and DRCC co-clustering on all datasets. It justifies that if a method uses more information such as relationships with other object types effectively, it can achieve improved results.  

Both STNMF and DRCC-Extended exploit similar amounts of information, yet STNMF gives lower performance compared to DRCC-extended. This is because STNMF reformulates the inter-relationships information as a symmetric matrix instead of clustering directly and simultaneously on the inter-relationship matrices as DRCC that was extended to include the multiple types using the concept of DiMMA. Hence, DRCC-Extended provides the second-best performance. 

MMNMF produces good performance on some datasets (e.g., D1, D2, D3) but fails to match other benchmarking methods on other datasets (e.g., D4, D5). This is due to its dependence (of consensus manifold learning and preserving process) on how original data points are sampled on the intrinsic manifold. Interestingly, RHCHME yields good performance on datasets D3, D4 and D5 compared to datasets D1 and D2. It aims to learn the complete and accurate intra-type relationships. However, it houses a complicated process of considering data lying on manifolds as well as data belonging to subspaces. Consequently, RHCHME often consumes a longer time to converge compared to its counterparts (Table \ref{tab:time}). 

It is interesting to note that, in some datasets such as D1 and D2, the performance of DRCC, DRCC-Extended and STNMF is inferior to NMF, whereas DiMMA always outperforms NMF. During clustering, DRCC, DRCC-Extended and STNMF incorporate the intra-manifold learning that relies on a $k$NN graph. This process may not produce the expected outcome as the difference between distances from a particular point to its nearest and its farthest points becomes non-existent in high-dimensional data \cite{Beyer1999}. However, DiMMA controls this by including inter-manifold learning.

\subsection{DiMMA without Inter-manifold}
We modified the objective function of DiMMA to exclude the inter-manifold component and tested how it affects low-rank representation learning. Table \ref{tab:DLcompare} shows that DiMMA consistently outperforms DiMMA-ExInter on all datasets which ascertains the effectiveness of the proposed objective function for the multi-aspect data.

\begin{figure}[t]
	\centering
	\includegraphics[width=0.7\linewidth]{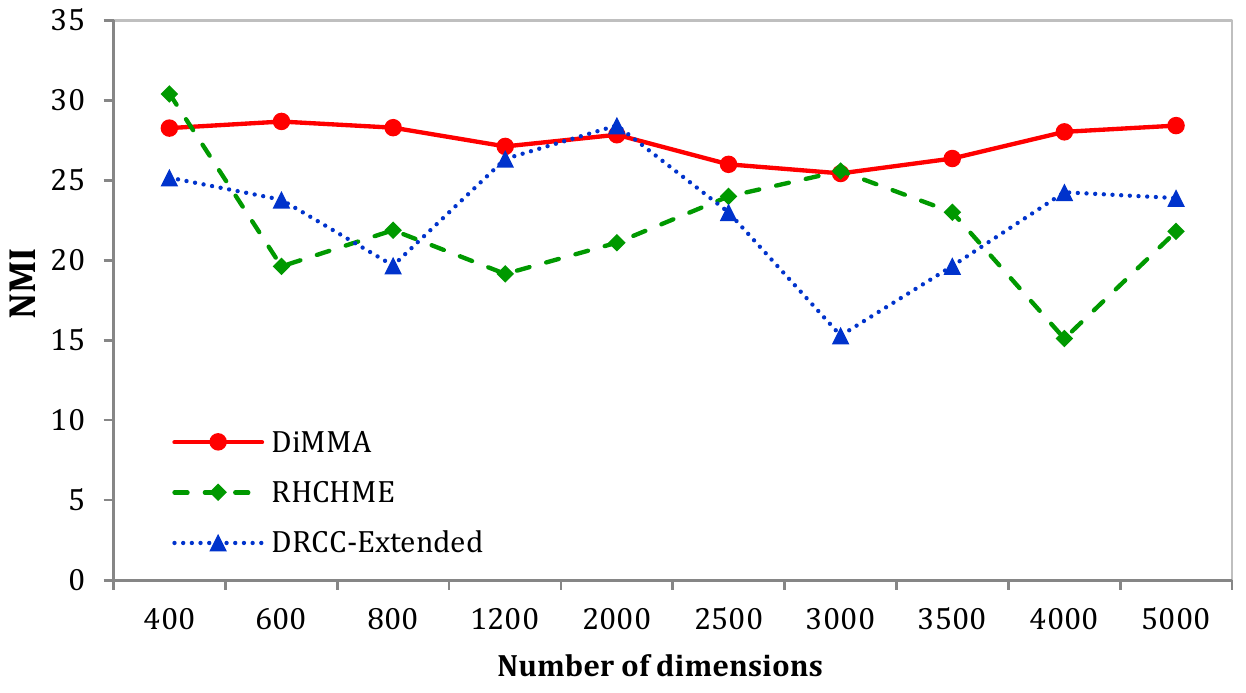}
	\caption{NMI of DiMMA with the increase in dimensionality.}
	\label{fig:dimen_inc}
\end{figure}
\begin{figure}[h]
	\centering
	\includegraphics[width=0.7\linewidth]{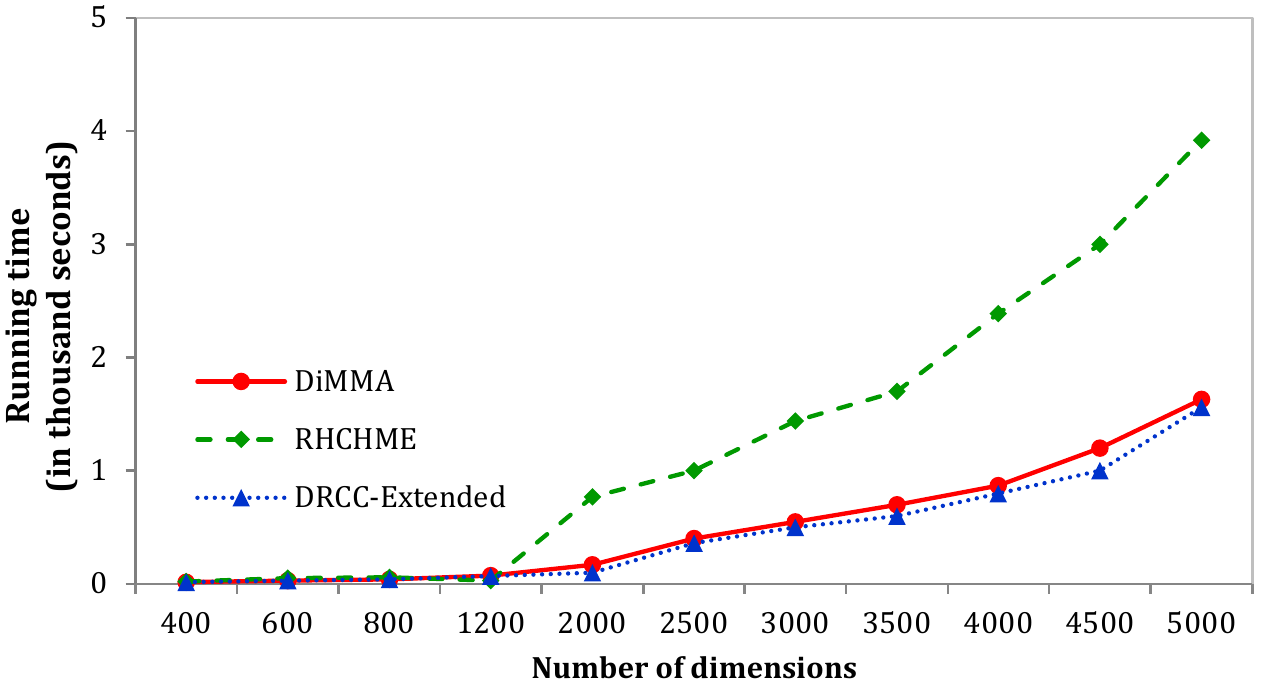}
	\caption{Time taken by DiMMA with the dimensionality increase}
	\label{fig:dimen_inc_time}
\end{figure}
\begin{figure}[t]
	\centering
	\includegraphics[width=0.7\linewidth]{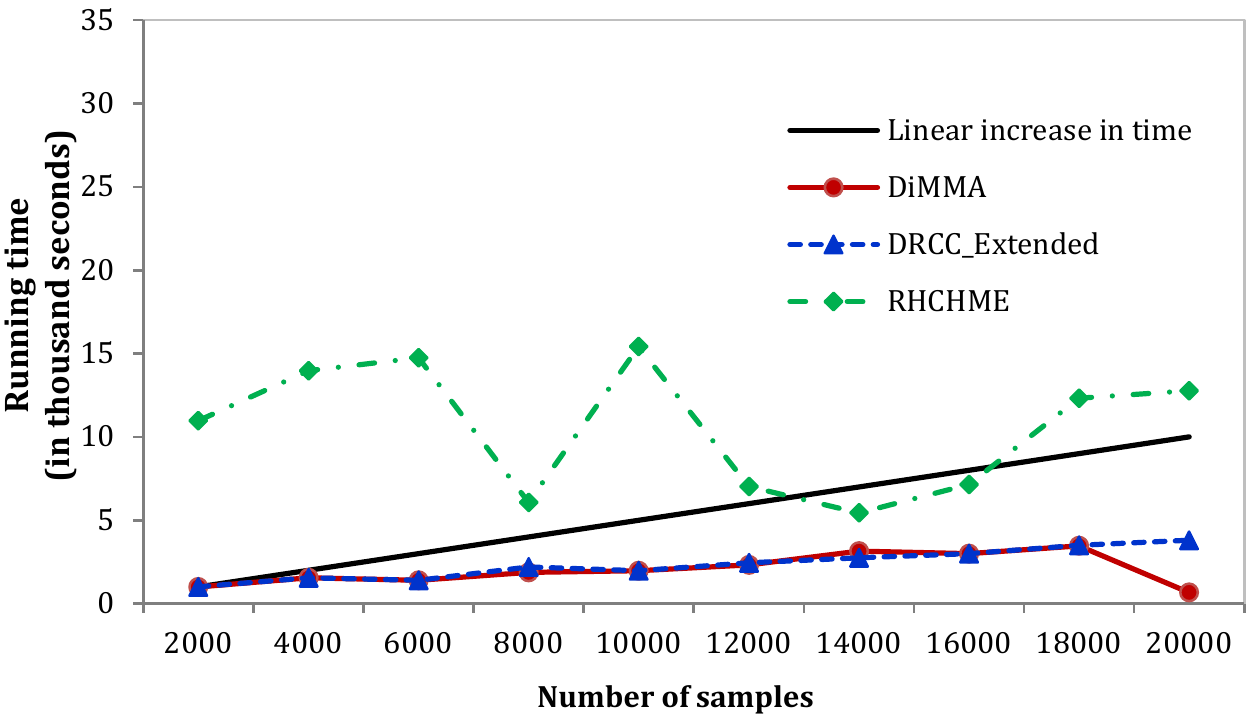}
	\caption{Scalability performance of DiMMA}
	\label{fig:perform_inc}
\end{figure}
\subsection{Comparison with deep learning-based Representation}
Since DiMMA attempts to learn a low-rank representation before conducting clustering, its performance is compared with deep-model-based methods. We experimented with the leading deep learning method DCCA \cite{DCCA:2013} that relies on a portion of data for training the deep neural network. We used a default setting for DCCA implementation as in \cite{DCCAE:2015}. We randomly selected $60\%$ data for training, $20\%$ for validation and use the rest $20\%$ data to test. Table \ref{tab:DLcompare} reports the clustering performance of all methods on the test datasets.  

The deep matrix factorization (DeepMVC) method \cite{DeepMVC} which builds a deep structure though matrix factorization framework is also used to show the effectiveness of using inter-manifold with DiMMA. There are four parameters in DeepMVC including the view weight distribution, the graph regularization, neighbourhood size and the layer size. We use default setting for layer size which is [100, 50] for all datasets as recommended in \cite{DeepMVC}. The graph regularization parameter is selected from the same range with DiMMA and the neighbourhood size is fixed as $5$. We turn off the view weight parameter since we aim to treat all views equally in this paper. 

\begin{figure*}[h]
\includegraphics[width=0.7\linewidth]{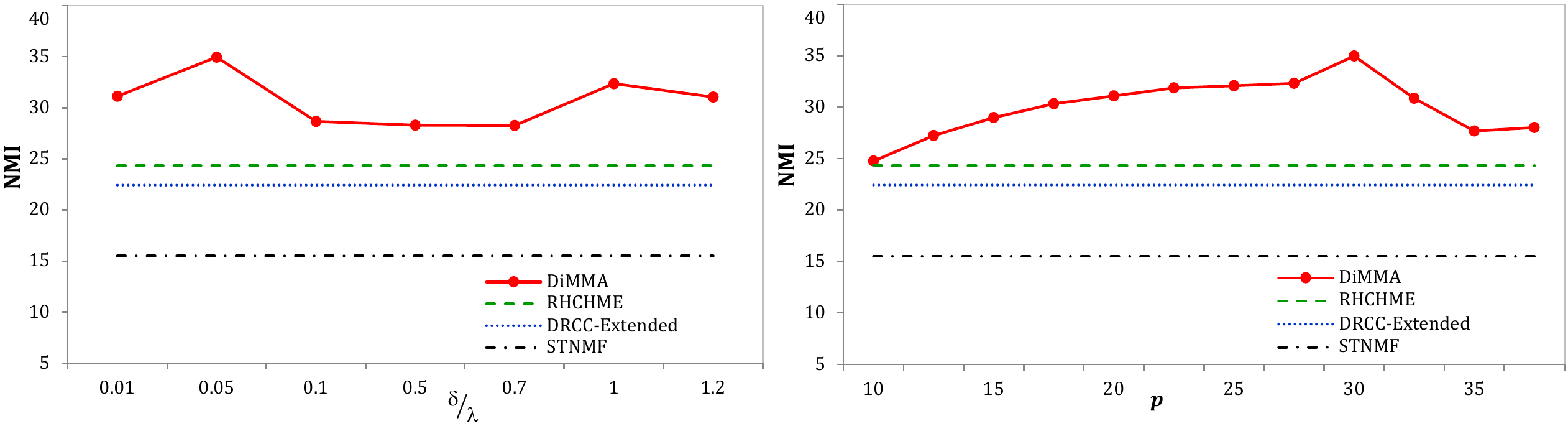}
\centering
\caption{Sensitivity analysis with intra- and inter-manifold learning regularization parameters ($\delta/\lambda$), and inter neighbourhood size $p$ on dataset MLR-1 (D1)}
\label{fig:D1_delta}
\end{figure*}
\begin{figure*}[t]
\includegraphics[width=0.7\linewidth]{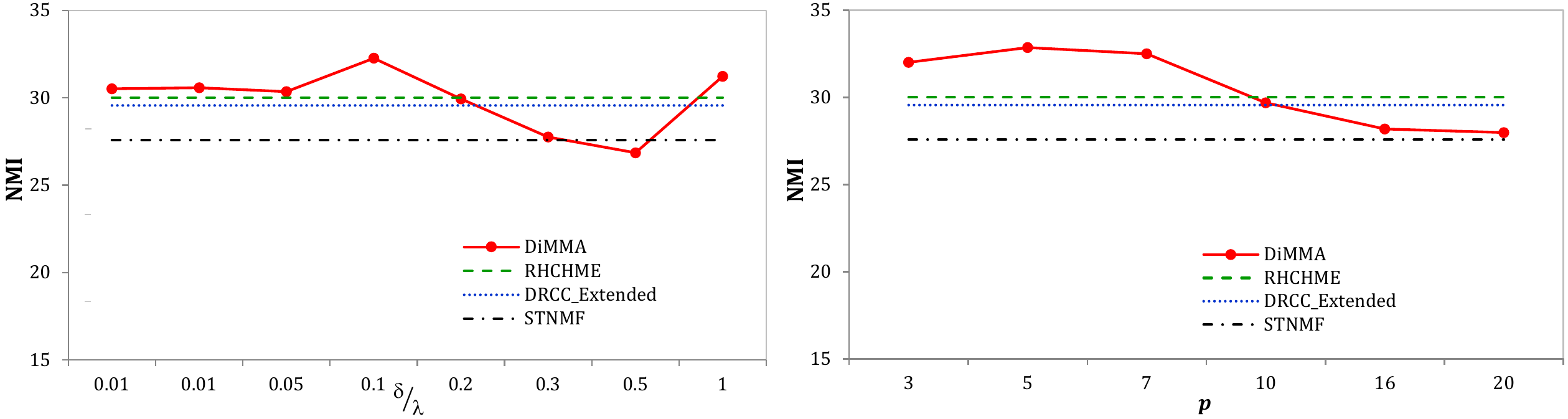}
\centering
\caption{Sensitivity analysis with intra- and inter-manifold learning regularization parameters ($\delta/\lambda$), and inter neighbourhood size $p$ on dataset Movie (D5)}
\label{fig:D5_delta}
\end{figure*}
Results show that DiMMA outperforms both DCCA and DeepMVC. A possible reason behind the poor performance of DCCA is the lack of a sufficient amount of training data. This emphasizes the need for NMF-based method to achieve an accurate clustering solution for high dimensional data in the unsupervised setting. DeepMVC though is based on a deep matrix factorization which is expected to return a good result, however, it may need the aid of advanced regularizations to deal with various dataset’s characteristics such as sparsity, cohesiveness of data samples in a cluster, etc. Investigating the effect of these characteristics of datasets on the clustering performance goes beyond the scope of this paper. However, it is worth mentioning that all benchmarking datasets except D6 are very sparse with the sparsity rate around 0.9. Dealing with sparse datasets is always a big challenge for state-of-the-art clustering methods. More importantly, cohesiveness of cluster solution on different datasets and methods in Table \ref{tab:cohes} can provide useful explanation for the poor performance of DeepMVC on datasets D1, D2 and D3. These datasets show a very low cohesiveness value in comparison to D4, D5 and D6. 

\subsection{Feature selection with clustering}
We claim that DiMMA supports feature selection during clustering (Lemma 2). We perform an experimental study to measure the cluster quality of DiMMA as compared to its NMF-based counterparts that support no such feature selection guidance. We propose cluster cohesiveness \cite{Cohesiveness:2000}\cite{Cohesiveness:2008} to evaluate feature selection methods applied in clustering. The cohesiveness of a cluster $c_i$ is calculated as, 
\begin{eqnarray}
Cohesiveness(c_i) = \frac{1}{|c_i|^2}\sum_{x, x' \in c_i} cosine(x,x')
\label{eq:cohes}
\end{eqnarray}
It reflects the average pairwise similarity between samples in each cluster, therefore, a higher cohesiveness shows better features selected during the learning process. The cohesiveness of a cluster solution $C$ is calculated as, 
\begin{equation}
Cohesiveness(C) = \sum_{i=1..c} \frac{Cohesiveness(c_i)}{|c_i|}
\end{equation}
where $c$ is the number of clusters in the cluster solution $C$. 

Results in Table \ref{tab:cohes} show that DiMMA achieves higher cohesiveness on all datasets. It confirms that better features have been selected during the learning process of DiMMA. 

\subsection{Time Complexity and Scalability of DiMMA}

DiMMA consumes substantially less time than MTRD methods of RHCHME and marginally better than STNMF and DRCC-Extended (Table \ref{tab:time}). These methods are based on reformulating input matrices to be a symmetric matrix thus require more running time, especially on large and high dimensional datasets. RHCHME always consumes the most time as a consequence of the complex process to learn data subspace. DiMMA consumes slightly more time than DRCC and MMNMF. However, the trade-off by obtaining a significantly improved accuracy for DiMMA justifies this.

We investigated the impact of the data size increase on the performances of DiMMA. We selected two benchmarking methods, DRCC-Extended and RHCHME, for this purpose. DRCC-Extended is the closest framework to DiMMA and RHCHME is the effective MTRD method based on the typical symmetric framework. We ignore NMF and DRCC in this experimental investigation since they are single or bi-type based methods. STNMF should have the same behaviour as RHCHME and MMNMF is a multi-view based method. 

We conducted experiments on a series of datasets that are extracted from the Reuters RCV1/RCV2 Multilingual dataset for this investigation. To check dimensionality effects, a dataset with 600 samples and 400 features on each view, i.e., English term, French term, German term and Italian term, was chosen as a starting point and then feature dimensions were increased systematically up to 5000. As can be seen from Figures \ref{fig:dimen_inc} and \ref{fig:dimen_inc_time}, DiMMA requires much less time than RHCHME and about the same time as DRCC-Extended to produce a high and stable result. DiMMA constructs a $p$NN graph $Z$ as in Eq. (\ref{eq:Zkl}) without needing to calculate similarity but using the available inter-relationship information as a distance approximation.

To check the effect of collection size, another dataset with $2000$ English documents was selected and the size was multiplied in each iteration (Figure \ref{fig:perform_inc}). The numbers of English, French, German and Italian terms are kept fixed at $5000$. DiMMA performs better than the linear time increase. 

\subsection{Parameters setting}
To enable a fair comparison between methods, we set the same value for all common parameters among methods. The intra-type neighbourhood size $k$ for constructing a $k$NN graph for each object type is fixed to $5$. Intra-type regularization parameter $\lambda$ is set to $10$ for datasets D1, D3, D5 and D6, and $1$ for D2 and D4. DiMMA has two additional parameters, the inter-regularization parameter $\delta$ and neighbourhood size $p$. Since DiMMA aims at incorporating both intra- and inter-manifold learning, the parameters for the intra- and inter-manifold can be set to the same value, treating intra- and inter-manifold learning equally. We set $\delta$ by using the ratio $\delta/\lambda$ from $0.01$ to around $1$ for each dataset. Similarly, a suitable $p$ value is set between $5$ to $30$. Figures \ref{fig:D1_delta} and \ref{fig:D5_delta} show that DiMMA can achieve the highest value when the right values have been chosen for these parameters. Since parameters $\delta$ and $p$ do not apply to benchmark methods, there is no change with the alterations of these parameters. However, these graphs show that DiMMA's performance stays above other benchmarks in most cases.

\section{Conclusion}
\label{sec:conclusion}
This paper approaches the problem of multi-aspect data clustering generically and provides a comprehensive analysis of factors that affect the problem such as ranking the features, relatedness between objects, data points' distances and geometric structures. It presents a novel method DiMMA that incorporates the diverse geometric structures of intra-type and inter-type relationships in MTRD and the geometric structures of samples and samples-feature in multi-view data. Extensive experiments show the effectiveness of DiMMA over the relevant state-of-the-art methods. DiMMA shows superior performance on both types of multi-aspect datasets by capturing the inter relationships between different object types. In future work, we will decompose the affinity matrices and form an overall framework for NMF clustering with graph regularization. We will also investigate combining an NMF-based method to a deep learning network-based model to enhance the use of the NMF framework in unsupervised learning.

\bibliographystyle{IEEEtran}
\bibliography{sigproc}
\vskip -3\baselineskip plus -1fil
\pagebreak
\numberwithin{equation}{section}
\section*{Appendix A: Arriving to Eq. (\ref{eq:P1_final}) from Eq. (\ref{eq:P1_allkl_1})}

Decomposing the last term from Eq. (\ref{eq:P1_allkl_1}) into two parts, we have the following equivalent equation,
\begin{equation}
\begin{split}
\min \big( &\sum_{h,l=1,h\neq l}^{m}Tr(G_{h}^{T}T_{hl}^{r}G_{h})+ \sum_{h,l=1,h\neq l}^{m}Tr(G_{l}^{T}T_{hl}^{c}G_{l}) \\
-2 &\sum_{1\leq h<l\leq m}Tr(G_{h}^{T}Z_{hl}G_{l}) - 2 \sum_{1\leq l<h\leq m}Tr(G_{h}^{T}Z_{hl}G_{l})\big)
\end{split}
\end{equation}

Changing the roles of $h$ and $l$ in terms $2$ and $4$ leads to the following equation:
\begin{equation}
\begin{split}
&\min \big( \sum_{h,l=1,h\neq l}^{m}Tr(G_{h}^{T}T_{hl}^{r}G_{h})+ \sum_{h,l=1,h\neq l}^{m}Tr(G_{h}^{T}T_{lh}^{c}G_{h}) \\
& -2 \sum_{1\leq h<l\leq m}Tr(G_{h}^{T}Z_{hl}G_{l}) - 2 \sum_{1\leq h<l\leq m}Tr(G_{l}^{T}Z_{lh}G_{h})\big)
\end{split}
\label{eq:transf1}
\end{equation}
\begin{equation}
\begin{split}
\Leftrightarrow &\min \big( \sum_{h,l=1,h\neq l}^{m}Tr(G_{h}^{T}T_{hl}^{r}G_{h})+ \sum_{h,l=1,h\neq l}^{m}Tr(G_{h}^{T}T_{lh}^{c}G_{h}) \\
& -2 \sum_{1\leq h<l\leq m}Tr(G_{h}^{T}Z_{hl}G_{l}) - 2 \sum_{1\leq h<l\leq m}Tr(G_{h}^{T}Z_{lh}^{T}G_{l})\big)
\end{split}
\label{eq:transf2}
\end{equation}

Eq. (\ref{eq:transf1}) can be written as Eq. (\ref{eq:transf2}) due to the property of $Trace$, $Tr(G_{l}^{T}Z_{lh}G_{h}) = Tr(G_{h}^{T}Z_{lh}^{T}G_{l})$.

Finally, we have Eq. (\ref{eq:P1_final}) by replacing $T_{h}$ and $Q_{hl}$ as defined in Eq. (\ref{eq:T_k}), (\ref{eq:Q_kl}) into Eq. (\ref{eq:transf2}).

\section*{Appendix B: Proof of Theorem 1}
To simplify the notations, we drop index $h$ in the notation and re-write Eq. (\ref{eq:L(G)}) as follows, 
\begin{equation}
\begin{split}
L(G) &=  Tr(G^{T}Q^{+}G - G^{T}Q^{-}G - G^{T}A^{+} \\
&+  G^{T}A^{-} + GB^{+}G^{T} - GB^{-}G^{T})
\end{split}
\end{equation}

To prove Theorem 1, we need to first prove $Z(G, G')$ is an auxiliary function of $L(G)$ and then prove that $Z(G,G')$ converges and its global minimum is the update rule of $G$. 

\textit{1. Prove $Z(G, G')$ is an auxiliary function of $L(G)$}

Using Lemma 2 we have,
\begin{equation}
Tr(G^{T}Q^{+}G) \leq \sum_{ij}\dfrac{(Q^{+}G')_{ij}G_{ij}^{2}}{G'_{ij}}
\label{eq:inequa1}
\end{equation}
\begin{equation}
Tr(GB^{+}G^{T}) \leq \sum_{ij}\dfrac{(G'B^{+})_{ij}G_{ij}^{2}}{G'_{ij}}
\label{eq:inequa2}
\end{equation}

In addition, we have $a \leq \dfrac{a^2 + b^2}{2b}, \forall a,b>0$, thus 
\begin{equation}
Tr(G^{T}A^{-})= \sum_{ij}A_{ij}^{-}G_{ij} \leq \sum_{ij}A_{ij}^{-}\dfrac{G_{ij}^{2} + G'^{2}_{ij}}{2G'_{ij}}
\label{eq:inequa3}
\end{equation}

We also have the inequality $z \geq 1 + \log z, \forall z > 0$. Therefore we have following inequalities
\begin{equation}
Tr(G^{T}A^{+}) \geq \sum_{ij}A^{+}_{ij}G'_{ij}(1 + \log \dfrac{G_{ij}}{G'_{ij}})
\label{eq:inequa4}
\end{equation}
\begin{equation}
Tr(G^{T}Q^{-}G) \geq \sum_{ijk}(Q^{-})_{jk}G'_{ji}G'_{ki}(1 + \log \dfrac{G_{ji}G_{ki}}{G'_{ji}G'_{ki}})
\label{eq:inequa5}
\end{equation}
\begin{equation}
Tr(GB^{-}G^{T}) \geq \sum_{ijk}B^{-}_{jk}G'_{ij}G'_{ik}(1 + \log \dfrac{G_{ij}G_{ik}}{G'_{ij}G'_{ik}})
\label{eq:inequa6}
\end{equation}

From Eq. (\ref{eq:inequa1}-\ref{eq:inequa6}) we have $L(G) \leq Z(G,G')$. Furthermore we have $L(G) = Z(G,G)$ is obviously. 

Thus $Z(G,G')$ can be called as an auxiliary function of $L(G)$ (Definition 2). 

\textit{2. Prove $Z(G,G')$ converges and its global minimum is the update rule of $G$}

We use Hessian matrix, and need to prove that the Hessian matrix of $Z(G,G')$ is a positive definite diagonal matrix. 

Taking the first derivative of $Z(G,G')$ on $G$, we have
\begin{equation}
\begin{split}
\frac{\partial Z(G, G')}{\partial G_{ij}}=&2\lambda\frac{(Q^+G')_{ij}G_{ij}}{G'_{ij}} - 2\lambda (Q^-G')_{ij}\frac{G'_{ij}}{G_{ij}}\\
-&2A^+_{ij}\frac{G'_{ij}}{G_{ij}}+2A^-_{ij}\frac{G_{ij}}{G'_{ij}}\\
+&2\frac{(G'B^+)_{ij}G_{ij}}{G'_{ij}}-2(G'B^-)_{ij}\frac{G'_{ij}}{G_{ij}}
\end{split}
\end{equation}
The Hessian matrix of $Z(G,G')$ containing the second derivatives as in Eq.  (\ref{eq:Hessian}) is a diagonal matrix with positive entries. 
\begin{equation}
\begin{split}
\frac{\partial^2 Z(G, G')}{\partial G_{ij}\partial G_{kl}}=&\delta_{ik}\delta_{jl}(2\lambda\frac{(Q^+G')_{ij}}{G'_{ij}}+2\lambda(Q^-G')_{ij}\frac{G'_{ij}}{G^2_{ij}}\\
+&2A^+_{ij}\frac{G'_{ij}}{G^2_{ij}}+2\frac{A^-_{ij}}{G'_{ij}}\\
+&2\frac{(G'B^+)_{ij}}{G'_{ij}}+2(G'B^-)_{ij}\frac{G'_{ij}}{F^2_{ij}})
\end{split}
\label{eq:Hessian}
\end{equation}
Thus $Z(G,G')$ is a convex function of $G$. Therefore, we can obtain the global minimum by setting $\partial Z(G, G')/\partial G_{ij} =0$. After some transformations and replacing index $h$, we can get Eq.  (\ref{eq:Gh}).
\end{document}